\documentclass[lettersize,journal]{IEEEtran}
\usepackage{amsmath,amsfonts}
\usepackage{algorithmic}
\usepackage{algorithm}
\usepackage{array}
\usepackage[caption=false,font=normalsize,labelfont=sf,textfont=sf]{subfig}
\usepackage{textcomp}
\usepackage{stfloats}
\usepackage{url}
\usepackage{verbatim}
\usepackage{graphicx}
\usepackage{cite}
\usepackage{amsmath}
\usepackage{microtype}
\usepackage{float}
\usepackage{booktabs}
\usepackage{soul}
\usepackage{url}
\usepackage{color}
\usepackage{makecell}
\usepackage{MnSymbol}
\usepackage{amsmath}

\begin{document}

\title{A Review and Roadmap of Deep Learning Causal
Discovery in Different Variable Paradigms}

\author{Hang Chen*, Keqing Du*, Xinyu Yang, Chenguang Li
        % <-this % stops a space
\thanks{Manuscript received September 13, 2022.}% <-this % stops a space

\thanks{The authors are with the Affective Computing and Intelligent
Analysis Laboratory, Xi'an JiaoTong University, Xi'an 710049,
China (e-mail:albert2123@stu.xjtu.edu.cn; dukeqing@stu.xjtu.edu.cn; yxyphd@mail.xjtu.edu.cn;
adgjl357159@stu.xjtu.edu.cn). * Hang Chen and Keqing Du contributed equally to this work. }}

\markboth{Journal of \LaTeX\ Class Files,~Vol.~14, No.~8, September~2022}%
{Shell \MakeLowercase{\textit{et al.}}: A Sample Article Using IEEEtran.cls for IEEE Journals}

\IEEEpubid{0000--0000/00\$00.00~\copyright~2022 IEEE}

\maketitle

\begin{abstract}
  Understanding causality helps to structure interventions to achieve specific goals and enables predictions under interventions. With the growing importance of learning causal relationships, causal discovery tasks have transitioned from using traditional methods to infer potential causal structures from observational data to the field of pattern recognition involved in deep learning. The rapid accumulation of massive data promotes the emergence of causal search methods with brilliant scalability. Existing summaries of causal discovery methods mainly focus on traditional methods based on constraints, scores and FCMs, there is a lack of perfect sorting and elaboration for deep learning-based methods, also lacking some considers and exploration of causal discovery methods from the perspective of variable paradigms. Therefore, we divide the possible causal discovery tasks into three types according to the variable paradigm and give the definitions of the three tasks respectively, define and instantiate the relevant datasets for each task and the final causal model constructed at the same time, then reviews the main existing causal discovery methods for different tasks. Finally, we propose some roadmaps from different perspectives for the current research gaps in the field of causal discovery and point out future research directions. 
\end{abstract}


\section{Introduction}
\IEEEPARstart{C}{ausality} is a relationship between an outcome and the treatment that cause it.\cite{pearl2009causality} It is ubiquitous in our lives involved in several fields such as statistics\cite{cox1992causality,heckman2022causality,eells1991probabilistic,berzuini2012causality}, economics\cite{hicks1980causality,heckman2008econometric}, computer science\cite{scholkopf2022causality,xu2020causality,meliou2010causality,raynal1996logical}, epidemiology\cite{vlontzos2022review,gottlieb2002relational,castro2020causality}, and psychology\cite{child1994causality,young2022development}. Take a common phenomenon in life, for example, many people wearing umbrellas because it is raining, or a student doing badly in an exam because he has not studied. This direct relationship between cause and effect is the simplest expression of causality. However, we need to be aware of the differences between statistical correlations and causality\cite{schield1995correlation,simon2017spurious}. For example, nylon socks and lung cancer teemed simultaneously in the last century, we can only conclude there was a correlation between them but not causality, owing to smoking also rose at this time. In recent years, the study of causality has become a crucial part of the field of artificial intelligence, thus overcoming some limitations of statistics-based machine learning\cite{boutaba2018comprehensive,smith2022real,guo2020survey}. Based on the directed acyclic graph (DAG)\cite{series2010guide,sauer2013use} structure and Bayesian models\cite{vowels2021d,gelman2014understanding}, it aims to learn the statistical relationship between two observed variables under the influence of another variable. Moreover, causality can generally be divided into two main aspects, causal discovery and causal effect inference\cite{spirtes2016causal,2018The}. The causal discovery\cite{eberhardt2017introduction,cooper2013causal} focuses on obtaining causal relationships from observed data and constructing structural causal models (SCMs)\cite{cinelli2019sensitivity,bongers2021foundations}, so that causal effect inference\cite{louizos2017causal,pearl2010causal,pearl2009causal} can estimate the change of a variable through the SCMs. Causal discovery is regarded as a necessary way, as well as a prerequisite for causal inference, has been attracting most attention in recent years.

\IEEEpubidadjcol

Causal discovery is the process of identifying causality starting with establishing a causal skeleton, further ending with a rigorous DAG usually called SCMs by related algorithms\cite{guo2020survey}. The causal skeleton\cite{ding2020reliable,2018The} refers to a completely undirected graph, all the pairwise variables are connected by undirected edges in it. Then, the causal algorithms used on the causal skeleton according to the statistical methods such as conditional constraint and independent component analysis\cite{stone2004independent}, the undirected edges are oriented and obtained SCMs where each directed edge represents the effect of one variable on another. Back in the early days of machine learning\cite{0Graphical,spirtes2000causation}, it proposed methods based on conditional constraints such as IC\cite{rosa2011inferring,chicharro2014algorithms}, SGS\cite{ramsey2012adjacency} and PC\cite{uhler2013geometry,spirtes2000causation, kalisch2007estimating}, and the score-based method GES\cite{chickering2002optimal} was came up with later on, these traditional approaches proposed correct causal assumptions and combined graph models to discover causality. Then, the methods based on Functional Causal Models (FCMs) including LiNGAM\cite{shimizu2006linear,shimizu2011directlingam}, and ANM\cite{hoyer2008nonlinear,peters2014causal} was put forward , further improving the computational efficiency and applicability of the model. These are the mainstream causal discovery methods, so there are many hybrid methods\cite{tsamardinos2006max,cai2018self,wong2002hybrid} and improved methods that combined their strengths.

Mentioned above are suitable for exploring causal relationships between the multiple endogenous variables with a certain number and value, and it's also the initial area of causal discovery studied. In light of the abundant research foundations, the causal discovery has gradually spread to the fields of pattern recognition\cite{mittal2021affect2mm,zhu2020cookgan}, such as image pattern recognition and text pattern recognition. Researchers found causality is also between different regions and parts in these endogenous binary-variable samples, such as in face recognition\cite{chen2021towards,oh2021causal}, fine-grained recognition\cite{rao2021counterfactual}, text emotion recognition\cite{sridhar2019estimating}, and other tasks\cite{egami2018make,zhang2020quantifying}. This kind of causal discovery method requires interpreting the correlation between the sample and label in traditional pattern recognition as the causal structure of each region or part of the recognizable sample, according to the researchers' prior knowledge or modeling needs. With the gradual diversification of causal discovery methods in this field, we consider if there exists another type of more complex variable. On the one hand in the perspective of a task, the overall achievement in static tasks such as recognition, classification, and segmentation has prompted researchers to explore dynamic sequences composed of a series of static tasks; on the other hand, in the perspective of models, the deepening of mainstream network models means simple tasks no longer reflect the gap between models, so there is a growing need for finer-grained labels and more interpretable research. 
\IEEEpubidadjcol
All these reasons promote the research field of causal discovery to go deeper into sequence tasks in 
the deep learning area.

In addition, the roadmap of causal discovery is the construction of USCM. According to the idea 
of existing methods, we propose three roadmaps, priori-based, sampling-based, and deterministic-based 
approaches. Causality is essentially a theory that considers potential causes\cite{imbens2020potential} beyond two variables. 
In terms of the original purpose for the creation of causal theory, if we just limiting it to a definite 
or semidefinite causal skeleton is not enough close to the realistic causality. As deep learning 
continues to evolve, USCM is the ultimate goal for causal theory to approach real-world causation. 
This will also drive us to address more causally relevant tasks, such as the process of constructing 
affective and knowledge products. Moreover, research based on interventions and counterfactuals can 
go further to try to reach the next stage of the AI field.

Overall, our contributions are as follows. First of all, we define the three types of tasks and 
illustrate their process; Secondly, we define the three types of variable datasets and compare their 
different characteristic; Thirdly, we define the three types of variable causal paradigm and analysis 
their processing of how to build up the SCMs; Finally, for the new challenge of USCM we propose some 
roadmaps to solve the problem of lack of causal discovery method caused by insufficient sampling.

The remaining sections of the paper are organized as follows. The second section defines definite task, 
MVD and DSCM, summarizes some common MVD and causal discovery methods under this paradigm. The third 
section defines semidefinite task, BVD and SSCM, collates BVD in different domains and the corresponding 
causal discovery methods. Likewise, we define undefinite task, IVD, and USCM, sums up the existing common 
datasets and associated tasks, further compares the similarities and differences between the three 
datasets and SCM in section \uppercase\expandafter{\romannumeral4}. In light of this, we analyze the 
current challenge and put forward the roadmaps in section \uppercase\expandafter{\romannumeral5}. The 
last section draws a general conclusion of this paper.

\section{Definite Task}
This section concentrates on the construction of DSCM based on the multi-variable dataset (MVD) in the 
definite task. In section \emph{A}, we define the definite task. Then, we define and instantiate MVD in 
section \emph{B}. In section \emph{C} we define the definite structural  causal model (DSCM). 
Besides, we sum up different types of causal discovery methods for DSCM, including constraint-based in 
section \emph{D}, score-based in section \emph{E}, the algorithms based on functional causal structure 
models in section \emph{F}, and the mixture of the above methods in section \emph{G}. Finally, we 
summarize these approaches in section \emph{H}. 

\subsection{Definition of Definite Task}
In general, the change of one variable often leads to multiple variables change, then will influence 
multiple variables to interact with each other and produce a series of changes. Therefore, exploring 
the pure causal relationship between pairwise variables often needs to exclude the interference of other 
variables\cite{pearl1988probabilistic}. SCMs use DAG\cite{glymour2019review,vanderweele2007directed} to 
describe causality, with nodes representing variables and edges representing the direct causal 
relationship. Then, although the variables in the real environment affect each other, if it is limited 
to a certain two variables, only the most closely related variables in the graph can be considered.

Let's consider two variables $V$ and $W$, variables $V$, $W$, $X_1$,$X_2$,\dots,$X_s$, $Z_1$,$Z_2$,\dots
,$Z_s$ are the most closely variables of $V$ and $W$, where $X_1$,$X_2$,\dots,$X_s$ are variables 
generally recognized to be related to $V$ and $W$ which can be sampled directly, $Z_1$,$Z_2$,\dots,$Z_s$ 
are variables not recognized to be related to $V$ and $W$ or difficult to sample directly. Therefore, 
the existing causal discovery tasks can be divided into three tasks according to the difficulty of the 
research and the method of solving the problem. The first task is also the most common and primitive 
task in causal discovery, aims to discover causal relationship between multiple definite variables. 
Thus, we can definite it as follows: 

\textbf{Definition 1 (Definite Task)} Explore the causality between $V$ and $W$ only under the influence 
of the variables $X_1$,$X_2$,\dots,$X_s$, $s=1,2,\dots,p$ is known and definite.

\begin{figure}
  \includegraphics[width=0.95\linewidth]{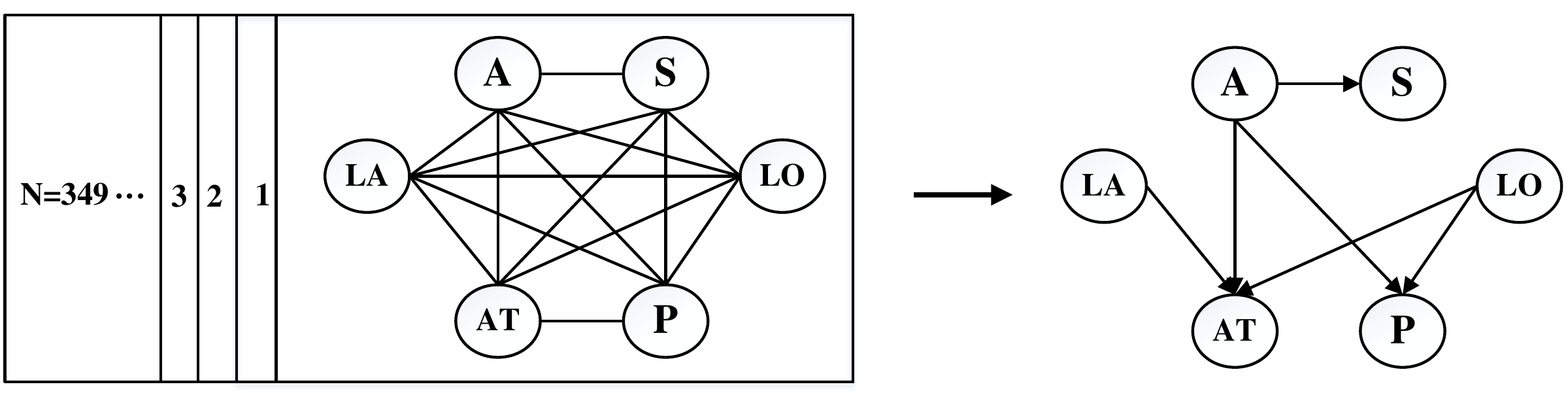}
  \caption{Take the DWD climate dataset\cite{dietrich2019temporal} for an xample to explore a 
  definite task. The dataset comes from 349 weather stations in Germany, including six variables: altitude (A), 
  latitude (LA), longitude (LO), sunshine hours (S), annual averages of temperature (AT) and precipitation (P). 
  The causal skeleton can be constructed directly from the datapoint of a dataset, and further construct the unique 
  DSCM.}
  \label{fig1}
\end{figure}

In this type of task, $V$, $W$ and $X_i$ are equivalent, and it's not limited to obtaining the causality 
between $V$ and $W$, but also all direct or indirect causal relationships between other two variables. 
This type of task applies to multi-variable datasets and aims to construct a definite causal 
structure model, the dataset and SCM we will define below. The process of definite task is shown in 
Figure~\ref{fig1}. In definite tasks, according to the known equivalent variables, we can obtain only 
one and the same causal skeleton, and further construct only one SCM through causal discovery methods. 

\subsection{Multi-Variable Dataset and Examples}
The dataset applied to definite task , have three or more equivalent and certain variables, often 
considered in statistical analysis for regression because of there is a causal relationship between 
its variables. Therefore, for definite task, we can construct multi-variable dataset contains $V$, 
$W$,$X_1$,$X_2$,\dots,$X_s$ as follows:

\textbf{Definition 2 (Multi-Variable Dataset)} A set is in form of {($X_{s+1}$,$X_{s+2}$,$X_1$,$X_2$,
\dots,$X_s$) $=$ ($v^{(j)}$,$w^{(j)}$,$x_1^{(j)}$,$x_2^{(j)}$,\dots,$x_s^{(j)}$), $j=1,2,\dots,k$ } is 
multi-variable dataset (MVD), and ($X_{s+1}$,$X_{s+2}$) = ($v^{(j)}$,$w^{(j)}$) represents $V$, $W$, 
and $X_i$ are equivalent.

\begin{table*}
\caption{ Some examples of multi-variable dataset.}
  \footnotesize
  \centering
  \begin{tabular}{|c|c|c|}
  %\resizebox{\linewidth}{!}{
    \hline
    \textbf{Data sets}&\textbf{Number of variables} & \textbf{Contents}\\
    \hline
    Erdos-Rényi Graphs & Customization & Number of probabilities based on custom 
    nodes $p = \frac{2e}{d^2}$ adding edges\\
    SynTReN\cite{van2006syntren} & Customization & Synthetic gene expression data \\
    Census Income KDD\cite{web:US,ristanoski2013discrimination} & 3 & The Census Income (KDD) 
    data set (US)\\
    Economic data\cite{iacoviello2019foreign} & 4 & Quarterly data for the United States 1965 - 2017\\
    Arrhythmia\cite{asuncion2007uci} & 4 & The Arrhythmia data set from the UCI Machine Learning 
    Repository concerns cardiac arrhythmia \\
    DWD climate\cite{dietrich2019temporal} & 6 & Data from 349 weather stations in Germany\\
    Archaeology\cite{huang2018generalized} & 8 & Archaeological data\\
    Soil Properties\cite{solly2014factors} & 8 & Biological root decomposition data \\
    AutoMPG\cite{asuncion2007uci} & 9 & Fuel consumption data for city car cycles\\
    Abalone\cite{nash1994population} & 9 & Conch abalone size data\\
    HK Stock Market\cite{huang2020causal} & 10 & Dividend-adjusted daily closing prices of 10 
    major stocks \\
    Sachs\cite{sachs2005causal} & 11 & Proteins and phospholipids in human cells\\
    Breast Cancer Wisconsim\cite{asuncion2007uci} & 11 & Breast cancer cell data\\
    CogUSA study\cite{tu2019causal} & 16 & Cognitive ageing data\\
    fMRI\cite{wang2003training} & 25 & The different voxel data in fMRI were organised into ROIs\\
    \hline
  \end{tabular}
  \label{sec2-tab1}
\end{table*}

In MVD, each variable is definite and each datapoint in the dataset contains a sample of 
variable X. Here are some examples as shown in Table~\ref{sec2-tab1}. The Erdos-Rényi model 
is a custom model, so that researchers can construct random graph models with n nodes according 
to the probability $p = \frac{2e}{d^2}$ in independently at the nodes $(i,j)$. SynTReN\cite{van2006syntren} 
acts as a network generator, it creates synthetic transcriptional regulatory networks and produces 
simulated gene expression data that approximates experimental data. In Census Income 
KDD\cite{web:US,ristanoski2013discrimination}, there is a causal relationship between the 
three variables, AAGE (age), AHRSPAY (wage per hour), and DIVVAL (dividends from stocks), and 
the DSCM can be constructed on them. Economic data\cite{iacoviello2019foreign} contains quarterly 
data on economic indicators for the US from 1965 - 2017 and can be used to investigate the causal 
relationships between GDP, inflation, economic growth and unemployment rates. 
Arrhythmia\cite{asuncion2007uci} is case record dataset from patients with arrhythmias, 
including four causally related variables: age, height, weight and heart rate, and is often used 
in regression prediction tasks to determine whether a patient will develop an arrhythmia. 
DWD climate\cite{dietrich2019temporal} comes from 349 weather stations in Germany, with no missing 
data, and its elevation, latitude, longitude and annual averages of sunshine hours, and temperature 
can be used to construct DSCM. Soil Properties\cite{solly2014factors} contain a number of variables 
related to biological root decomposition, including soil properties such as clay content, soil organic 
carbon content, and soil moisture measured in both grassland and forest environments, in addition to 
decomposition rates. AutoMPG\cite{asuncion2007uci} as a common dataset for causal discovery 
experiments, it contains nine vehicle properties and their fuel consumption. Abalone\cite{nash1994population} 
is a morphological dataset for different conch abalone, including 9 variables, the data 
on their sex, length, diameter, height, whole weight, shucked weight, viscera weight, shell weight.

HK Stock Market\cite{huang2020causal} is the dividend-adjusted daily closing prices of 10 different 
stocks in Hong Kong from 10 September 2006 to 8 September 2010, the 10 stocks are calculated by 4 
different classification indices. There is also a causal relationship between the closing prices of 
different stocks due to inter-industry causality and measurement indices. Sachs\cite{sachs2005causal} 
is a causal signal dataset of proteins and phospholipids in human cells, it reflects the relationship 
between stimulation or inhibition of cells by different signals, and its variables are 11 chemical 
signals. The deep learning-based causal discovery methods SAM and the Augmented Lagrangian Method 
(ALM) with constrained optimization to solve the causal structure are all based on this dataset to 
verify the effectiveness. Breast Cancer Wisconsim dataset\cite{asuncion2007uci} was developed to 
explore the cause of cancer, it contains 11 variables about different types of cells associated 
with breast cancer, including cellular protein types, cellular telomeres and so on. Similarly, the 
CogUSA study\cite{tu2019causal} included 16 variables such as age, gender and BMI in order to 
investigate the causes of aging. fMRI\cite{wang2003training} organized the different voxel data 
into regions of interest (ROIs) according to functional magnetic resonance data including anatomical 
images and obtained 25 ROIs to analyze the causal relationships between the different ROIs.

\subsection{Definite Structural  Causal Model}
MVD is fundamental to obtain the causal skeleton between variables, and if we know just a few of the 
points, we can accurately deduce the rest according to SCM, it is also the final goal of definite 
task. So we put forward the definition of Definite structural causal model according to 
definite task and MVD:

\textbf{Definition 2 (Definite Structural  Causal Model)} Let $\mathcal{V}$ = ($V$,$W$,$X_1$,
$X_2$,\dots,$X_s$) is the set of variables in MVD, it forms the directed acyclic graph $\mathcal{G}$ = 
($\mathcal{V}$, $\mathcal{E}$) is definite structural causal model (DSCM).

In definite paradigm, according to the samples in MVD contain same definite variables, 
we can construct causal skeleton directly whose nodes corresponding to the variables in MVD. The 
causal skeleton in definite task is the same and only one, so that it produces only one structural causal model named DSCM. Moreover, we need some causal discovery methods to construct DSCM, 
we will propose them below.

\subsection{Constraint-based Algorithms}
Constraint-based causal discovery methods rely on statistical tests of conditional independence\cite{baba2004partial,prakasa2009conditional}, 
they are easy to understand and widely used. These methods usually under the assumptions of causal 
Markov property\cite{lauritzen2000causal} and faithfulness\cite{weinberger2018faithfulness,sadeghi2017faithfulness}, also 
assuming there are no unobserved confounding variables, so they efficiently search to obtain a Markov 
equivalence class\cite{andersson1997characterization} of graphs, that is a set of causal structures 
that satisfy the same conditional independence. Most constraint-based methods first estimate the 
possible skeletons also named undirected edges, then determine the collider (V-structure)\cite{cole2010illustrating} to obtain the CPDAG\cite{kalisch2010pcalg}, and then determine the direction of 
the other edges as far as possible.

\textbf{IC Algorithm (Inductive Causation algorithm):} The IC algorithm\cite{pearl2000models} works in two major steps. First, for each pair of variables $A$ and $B$ within the set of nodes, it finds the separation set $S_{AB}$ of the two variables, with $S_{AB}$ as the condition when $A$ and $B$ are independent of each other. If there is no separation $S_{AB}$ between $A$ and $B$, then $A$ and $B$ are always related, so that undirected edges are added to $A$ and $B$. A partially undirected graph is obtained after all two-way judgements. Then for each pair of nodes $A$ and $B$ that are not adjacent and have a common neighboring node $C$, its structure is in form of $A$-$C$-$B$, check whether $C$ $\in$ $S_{AB}$  and if so, proceed to retrieve the next pair, if not, this means $C$ as a condition does not make $A$ and $B$ independent, so it is considered a V-structure (collider): $A$ $\rightarrow$  $B$ $\leftarrow$  $C$. After determining all collision structures, a partially directed graph is obtained. The process of determining the orientation of the remaining undirected edges follows two conditions according to the rules: the orientation cannot create new collision structures and the orientation cannot generate directed loops.

There is also an IC* algorithm based on the IC algorithm, it has the same first two steps as the IC 
algorithm. In orienting the other remaining undirected edges of the partially directed graph, IC* 
identifies the remaining undirected edges as true causal relationships, potential causal relationships
, spurious correlations, and relationships that remain undetermined, giving explicit definitions of 
potential and true causal relationships. In all these definitions, the criterion for a causal 
relationship between two variables $X$ and $Y$ would require the third variable $Z$ exhibit a 
particular pattern of dependence on $X$ and $Y$, coinciding with the nature of the causal statement 
as the behavior of $X$ and $Y$ under the influence of the third variable. The difference only lies in 
the variable $Z$, as a virtual control, must be identifiable and knowable within the data itself. 
In contrast to the IC algorithm, the IC* setting releases the causal adequacy assumption and has no 
requirement for unobserved potentially confounding variables.

Exhausting the test of independence between all nodes is a very large search problem, it is also the 
disadvantage of this type of approach, and the amount of computation is unacceptable when the number 
of nodes is large. For this reason, there are also improved algorithms (such as the PC algorithm)  
perform the test by gradually increasing the number of nodes.

\textbf{PC Algorithm (Peter-Clark algorithm):} The PC algorithm\cite{spirtes2000causation, kalisch2007estimating} reduces unnecessary independence tests and searches compared to the IC algorithm and the SGS algorithm, the predecessor of the PC algorithm. PC algorithm can also be roughly divided into two steps. First, construct an undirected complete graph, and respectively retrieve whether there are pairs of variables A and B in the graph when $n = 0, 1, 2, \dots$ are conditioned on other $n$ variables. If satisfied, delete all undirected edges between two nodes like $A$ and $B$, add these n variables to the separation set $S_{AB}$ of $A$ and $B$, update the undirected graph, and increment $n$ until no two variables are in the When conditional on the remaining $n$ variables, it is shown as d-separation in the graph. After obtaining the partially directed graph with the irrelevant edges removed, find a three-vertex structure similar to $A-C-B$. If $C$ is not in $S_{AB}$, then the v-structure can be determined, and the remaining undirected edges can be determined based on other rules.

The difference between the PC algorithm and its predecessor, the SGS algorithm, is that some edges 
are removed from a completely undirected graph. Let $V$ be the set of points of all vertices, and for 
each pair of vertices $A$ and $B$, remove the edge between them from the completely undirected graph 
if there exists a subset $S$ of $V\setminus \{A, B\} $ such that $A$ and $B$ are d-separation in a 
given $S$. This search method of SGS is more stable compared to the PC algorithm, but its complexity 
is higher, and for each pair of variables adjacent to each other in the graph $\mathcal{G}$, we 
search for all remaining possible subsets of variables, which is an exponential search. In contrast, 
the PC algorithm is much less computationally intensive, let $k$ be the maximum degree of any vertex 
and let $n$ be the number of vertices. The worst-case time complexity can be limited 
to $2\binom{n}{2}\sum_{i = 0}^{k}\binom{n-1}{i} $, which converges 
to $\frac{n^2{(n-1)}^{k-1}}{{(k-1)}!}$.

\textbf{FCI Algorithm (Fast Causal Inference Algorithm):} Also based on causal Markov property and faithfulness, the FCI algorithm\cite{colombo2012learning, zhang2008completeness, shen2020challenges} is a generalization of the PC algorithm. However, unlike the three methods mentioned above, the generality of FCI lies in the fact that FCI can be used in the presence of unobserved confounding variables (i.e violating the causal sufficiency assumption), and its results are proven to be asymptotically correct. 

Firstly, construct a complete graph consisting of undirected edges. For vertices $X_i$ and $X_j$, 
given $X_i $ conditioned on the set $M$ of adjacency subsets, do a conditional independence test 
on the two vertices, if the conditional independence relation holds, remove the edges between them 
and save the corresponding neighbouring subset $M$ to the separating sets $Sepset(X_i, X_j)$ 
and $Sepset(X_j, X_i)$. The skeleton derived in the first step is the superset of the final skeleton. 
For an open triple $\left\langle {X_i,X_j,X_k}\right\rangle$, if $X_j$ is not in  $Sepset(X_i, X_k)$ 
and  $Sepset(X_k, X_i)$, then orient this triple into a v-structure. For each vertex $X_i$ in $V$, 
find its $pds(X_i,\cdot )$. Then for all $X_i$ the adjacency vertices  $X_j$, test whether 
there exists ${X_i} \upmodels {X_j}|{(Y \bigcup S)} $, where $Y\subseteq psd{(X_i,\cdot)}\setminus \{X_i,X_j\}$. If it exists then remove the edge between the two vertices and save the corresponding adjoint subset $M$ to the separating sets $Sepset(X_k, X_i)$ and $Sepset(X_i, X_k)$.This breaks the skeleton of the graph and requires a reorientation of the v-structure. So finally change all edges back to undirected edges. Repeat for all triples $\left\langle {X_i,X_j,X_k}\right\rangle$ are retrieved and the v-structure is reoriented. The remaining undirected edges are given as many edges as possible to orient according to the orientation criterion in  rules.

The FCI algorithm is a method for inferring the causal relationship between observable variables with latent variables and selected variables. In light of the PC algorithm, the conditional independence test is carried out on the two vertices of the existing edge, it is more universal and avoids the problems arising from the existence of unobservable latent variables and selection variables. It has the disadvantage of high time complexity. There also exists a variant of FCI, RFCI\cite{colombo2012learning}, which speeds up the algorithm by sacrificing less information.

\textbf{CD-NOD Algorithm (Constraint-based Causal Discovery from Heterogeneous/Nonstationary Data):} In real practice, mechanisms or parameters associated with causal models may change over data sets or over time, so some causal links in the structure may disappear or appear in certain domains or over time. CD-NOD\cite{zhang2017causal} is a constraint-based causal discovery approach from heterogeneous/Nonstationary data to efficiently identify variables with local mechanisms of change and use the information carried by distribution shifts to determine the causal direction.

All in all, constraint-based methods all follow Markov property and faithfulness, and their output is a causal graph with partially marked causal edges (PDAG), i.e., Markov equivalence classes, a set of causal structures that satisfy the same conditional independence, and any directed graph that satisfies the Markov equivalence classes satisfies the conditional independence set of the data. Therefore, the existence of DAGs that cannot be distinguished from Markov equivalence classes is also a general problem with constraint-based methods. These methods involve conditional independence tests, which would be difficult to implement if the form of the dependencies were unknown. It has the advantage that it is universally applicable, but also has the additional disadvantage that faithfulness is a strong assumption and it may require a very large sample size to get a good conditional independence test, due to faithfulness requires a large sample to be confirmed.

\subsection{Score-based Algorithms}
Traditional methods of causal discovery include, in addition to constraint-based methods, score-based methods, whose goal is to find causal structure by optimizing a suitably defined score function. For the disadvantage that faithfulness needs to be confirmed with large samples, the score-based methods will dilute the faithfulness assumption by applying a goodness-of-fit test instead of a conditional independence test. Similarly, under the causal adequacy assumption, these methods will maximize the score $S$ of a causal graph $\mathcal{G}$  given the data to find the optimal graph $\mathcal{G}$.

\textbf{GES Algorithm (Greedy Equivalence Search):} Unlike PC, SGS, and FCI, starting with an undirected complete graph to build the causal skeleton. GES\cite{chickering2002optimal,hauser2012characterization} starts with an empty graph, adding the edges currently needed and then eliminating unnecessary edges in the causal structure. At each step of the algorithm, when it is decided to add a directed edge to the graph will increase the fit as measured by the score function, the edge that best improves the fit is added, performing the \textbf{\emph{Insert operator}}. The resulting model is then mapped to the appropriate Markov equivalence class, and the update is repeated to continue the process. When the score can no longer be improved, the GES algorithm searches edge by edge, and if deleting any edge maximizes the score, it removes that edge, executing the \textbf{\emph{Delete operator}}, until deleting any edge can no longer increase the score function fit value. 

Scoring functions are often chosen from Bayesian scoring functions or information theory-based functions such as Bayesian information criterion (BIC)\cite{burnham2004multimodel,neath2012bayesian}, BDeu scoring criterion\cite{suzuki2017theoretical,liu2012empirical}, Generalized score~\cite{huang2018generalized}, K2 scoring\cite{kayaalp2012bayesian,chen2008improving}, MDL scoring\cite{carvalho2009scoring,jiang2011learning}, posterior scoring of graphs given data, and so on.

GES is a method for learning Bayesian networks from data using greedy search. Its use of the equivalence class of the graph as a search state can make the scoring of operators in the method more efficient, with the advantage of avoiding the difficulties that independence tests may have, and the disadvantage of high space complexity and low operational efficiency, and its inability to be applied in the presence of unknown confounders.

\textbf{fGES Algorithm (fast Greedy Equivalence Search):} fGES\cite{ramsey2017million,scheines2016measurement} adds two improvements to GES, parallelization and reorganization of the cache, for discovering directed acyclic graphs over random variables from sample values. fGES radically improves the running time of the algorithm.

Like constraint-based methods, score-based methods suffer from the problem of indistinguishability of Markov equivalence classes. This type of methods avoids the possible problems of independence tests, although it's not as simple to illustrate as the constraint-based methods. In the large sample limit, both algorithms converge on the same Markov equivalence class under almost identical assumptions.

\begin{figure}
  \includegraphics[width=0.95\linewidth]{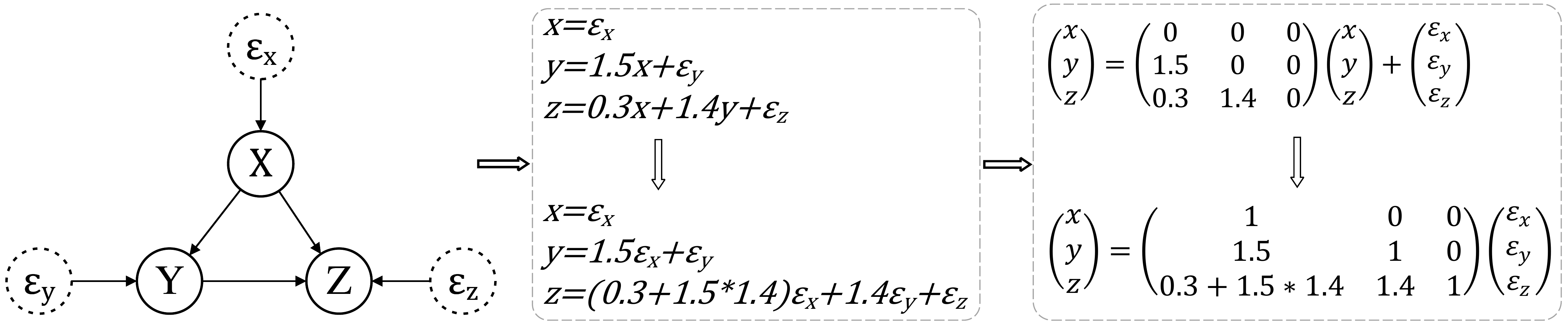}
  \caption{Three representations of LiNGAM, graphical model representation, equation representation and matrix representation.}
  \label{lingam}
\end{figure}

\subsection{Algorithms based on Functional Causal Models (FCMs)}
Unlike the previous two methods, FCMs effectively avoids problems such as the inseparability of the MEC and the need for large samples to confirm causal faithfulness. FCMs can obtain the entire causal graph under certain constraints on the functional class of the causal mechanism by exploiting the asymmetry between the causal and anti-causal directions. For continuous variables, the FCM represents the influence variable $Y$ as a function of the direct cause $X$ and some (unobservable) noise term $\mathcal{E} $, $Y = f (X, \mathcal{E} )$, where $\mathcal{E} $ is independent of $X$. Due to the constrained functions, the conditional independence between the noise $\mathcal{E} $ and the cause $X$ applies only to the true causal direction and not to the wrong causal direction, so that the causal relationship between $X$ and $Y$ the causal direction is identifiable.

\textbf{LiNGAM (Linear Non-Gaussian Acyclic Model):} LiNGAM\cite{shimizu2006linear} is a concrete causal relationships model based on a Bayesian network. Its assumptions : (1) the data generation process is linear, (2) there are no unobserved confounders, and (3) the disturbance terms have a non-Gaussian distribution with non-zero variance and are independent of each other. Its modeling idea is shown in Figure~\ref{lingam}.
%暂时认为图像没出现是因为没编到该出现图像的位置

The LiNGAM model can be written as $x = Bx + \varepsilon $; , $x$,$\mathcal{B}$ and $\varepsilon $ respectively denote the vector of variables, the adjacency matrix of the causal map and the noise vector. $x$ and the columns of $\mathcal{B}$ are respectively sorted according to the causal order $k(i)$ of each variable. Varying the equation yields $(I - B)x = \varepsilon $, so that $A = {(I - B)}^{-1}$ and $x = A\varepsilon $ is a pure equation, since $B$ is a strict lower triangular matrix, then $A$ is a non-strict lower triangular matrix. For $x = A\varepsilon $, we can use ICA\cite{stone2004independent} method to find theoretically $W = A^(-1)$ for a unique arrangement of rows such that it yields a matrix that contains no zeros on the main diagonal $\widetilde{W}$. But in practice a smaller estimation error would result in all elements of W being non-zero, so in practice one needs to find the alignment that makes $\sum_{i}\frac{1}{\left\lVert \widetilde{W}_{ii} \right\rVert }$ the smallest possible arrangement. By dividing $\widetilde{W}$ each row of the matrix by its corresponding diagonal element to obtain a new matrix $\widetilde{W}'$, with all elements of its diagonal being 1. Then, the estimate of $B$ can be calculated from $\hat{B} = I - \widetilde{W}'$, and calculate the estimate of $B$ as $\hat{B} $; finally, to find the causal order, look for $\hat{B}$ the substitution matrix $P$ , thus generating a matrix $\widetilde{B} = P\widetilde{B}P^T $ to get it as close as possible to the strict lower triangular shape, this can be verified by $\sum_{i\leq j}\widetilde{B}^2_{ij}$. Thus, the adjacency matrix $B$ of the causal diagram can be obtained, leading to the structural  causal model.

ICA estimates are usually according to the optimization of non-quadratic (possibly non-convex) functions, and the algorithm may fall into local minima. As a result, different estimates of $W$ may be obtained for different random initial points used in the optimization algorithm. However, typical ICA algorithms are relatively stable when the model holds and unstable when the model does not hold. Computational solutions to this problem need to be based on re-running the ICA estimates with different initial points. Therefore, LiNGAM may have a defect of local convergence, making the solution result often locally optimal rather than globally optimal.

\textbf{ANM (Additive Noise Models):} LiNGAM is able to identify causal structures thanks to the non-Gaussian nature of the noise terms which are non-symmetrical, but in reality, many causal relationships are more or less non-linear, and ANM\cite{hoyer2008nonlinear,peters2014causal} demonstrates non-linearity can also break the symmetry between variables, thus allowing us to identify the direction of causality between variables. ANM is based on two assumptions: (1) the observed effect $Y$ can be expressed as a function of the cause $X$ and additive noise $\varepsilon $ as a function of the model, $Y = f(X) + \varepsilon $, (2) the cause $X$ and additive noise $\varepsilon $ are independent.

If $f(\cdot )$ is a linear function and the noise is non-Gaussian distributed, the ANM works in the same way as LiNGAM . The model is learned by regression in two directions and testing for independence between the hypothesized cause and noise (residuals) in each direction, with the decision rule being to choose the less correlated direction as the true causal direction. For the two input variables, if $X$ and $Y$ are statistically correlated, and thus test if $Y = f(X) + \varepsilon $ matches the data, the corresponding residual is calculated $\varepsilon  = Y - f(X)$ and test for $\varepsilon $ and independence between $X$; similarly, if $X = f(Y) + \varepsilon $ matches the data, then calculate residual $\varepsilon  = X - f(Y)$ and test for $\varepsilon $  and $Y$; if neither of the two directions fits the data, the model cannot be fitted. Repeat the above steps until each pair of variables has been tested.

The ANM algorithm assumes that only one direction fits the model and cannot handle the linear Gaussian case, as the data can fit the model in both directions, so the asymmetry between cause and effect disappears. The improved ANM can be extended to the linear Gaussian case, and the improved model is also more effective in the multivariate case.

\textbf{PNL (post-nonlinear model):} In PNL\cite{zhang2012identifiability,zhang2015estimation}, the effect $Y$ is a nonlinear transformation of the cause $X$ plus some internal additional noise and then an external nonlinear transformation, denoted as $Y = f(g(X) + \varepsilon )$, whose residuals can be calculated $\varepsilon  = f^{-1}(Y) - g(X)$. If the two variables x and y are statistically correlated, test if $Y = f(g(X) + \varepsilon )$ fits the data, $\varepsilon  =f^{-1}(Y) - g(X)$ and test for the independence between $X$ and $\varepsilon $; similarly if $X = f(g(Y) + \varepsilon )$ fits, then $\varepsilon  = f^{-1}(X) - g(Y)$ and independence between $Y$ and $\varepsilon $; if neither of the two two directions fits the data, then the model cannot be fitted. Repeat the above steps until each pair of variables has been tested.

The estimation process of PNL assumes only one direction fits the model, its causal model takes into account the effects of cause, noise effects, and possible sensor or measurement distortions in the observed variables. The form of PNL is very general and its identifiability is demonstrated, but PNL is sensitive to the assumed noise distribution and a Bayesian inference-based PNL causal model estimation method allows for automatic model selection and provides a flexible model of the noise distribution, it can effectively address this issue.

\textbf{IGCI (Information Geometry Causal Inference Model):} The IGCI model\cite{janzing2012information} is based on the assumption, the marginal distribution $P(X)$ and the conditional distribution $P(Y | X)$ are independent of each other in a particular way if $X $ is the cause of $Y$. An applicable and explicit form of reference measurement is the entropy-based IGCI or the slope-based IGCI.

\textbf{FOM (Fourth-order moment Model):} The existing algorithms based on functional causal models almost always assume the data are homoscedastic; in heteroskedastic data, the assumption of independence between noise and cause variables no longer holds, and the previously described methods in light of this assumption are unable to identify the direction of causality. The assumption of homoskedasticity noise is usually unrealistic in real-world applications because of the ubiquity of unobserved confounding factors. The FOM\cite{cai2020fom} method based on the fourth-order moment asymmetric measure can solve the causal identification of heteroskedasticity data by setting its functional causal model $Y = f(X) + \alpha \varepsilon $,where the fluctuation factor $\alpha $ is used to represent the heteroskedasticity of the data.  The correct causal direction can be obtained by comparing the heteroskedasticity in different pairwise variable directions.

As indicated previously, in contrast to traditional causal discovery methods, FCMs represents the outcome as a function of the direct cause and an independent noise term, and suggests some pathways to address the assumption of no confounders. On the other hand, many tests for conditional independence make certain assumptions about the data distribution, functional form or additive noise in practice. When these assumptions are not met, the conditional independence test may fail. Therefore, the FCMs also do not restrict the values to the deterministic variable data set presented in this paper, and its applicability is much broader.

\subsection{Mixed methods}
\textbf{GFCI (GES \& FCI):} It is a method that combines GES and FCI, uses GES to find the hypergraph of the skeleton and FCI to prune the hypergraph of the skeleton and find the orientation. The GFCI algorithm\cite{jabbari2020instance,glymour2019review} has also been shown to be more accurate than the original FCI algorithm in many simulation experiments.

\textbf{MMHC (Max-Min Hill-Climbing) Algorithm:} The MMHC algorithm\cite{tsamardinos2006max,wang2018analysis} is extremely scalable for thousands of variables. First, it learns the skeleton of the causal graph using the MaxMin Parents and Children (MMPC) algorithm, which is similar to the constraint-based algorithm. Then, edges are localised using a Bayesian scoring hill-climbing search method similar to the score-based algorithm.

\textbf{SELF (Structural Equational Likelihood Framework):} Most of the previous studies have adopted two distinct methods under the Bayesian network framework, namely global likelihood maximization and marginal distribution local complexity analysis. For example, traditional causal discovery methods carry out causal discovery from a global perspective, while FCMs carry out causal discovery from a local perspective. SELF\cite{cai2018self} combines these methods to form a new global optimization model with local statistical significance. The noise is injected into the variables independently on the graphical model, given a set of structural equations corresponding to the variable generation process, the observed distribution of the variables is completely determined by the distribution of the noise. SELF focuses on noise estimation, which maximizes the global likelihood of the entire Bayesian network while maintaining the local statistical independence between noise and causal variables, taking into account both perspectives.

\textbf{SADA (Scalable Causation Discovery Algorithm):} SADA\cite{wong2002hybrid,raghu2018comparison} is a hybrid method combined constraint-based and FCMs. It can be used to solve the false discovery rate control problem on high-dimensional data. The causality discovery problem through SADA can be decomposed into sub-problems and solved using recursive methods to improve the accuracy of the algorithm.

\subsection{Summary}
In this section, we definite the initial causal discovery task as definite task, dataset MVD used in this task, and the final model DSCM to show the causal relationship in known multiple variables. As indicated previously, there are different types of causal methods we can use to construct DSCM. The constraint-based and score-based methods are relatively easy to understand, but they are limited by assumptions that lead to some problems such as MEC inseparability, the need for large samples to prove faithfulness, and the inability to deal with potential confounders. Therefore, the methods based on FCMs are universal to apply through formalizing causality in terms of matrices and introduced the concept of exogenous variables also named noise term for the first time. These methods had an assumption, such as the independence between data and non-Gaussian properties of exogenous variables. They can not only solve the above limitations but also avoid other problems that may be encountered in conditional correlation or score functions. The benefit of a hybrid approach is that it combines the strengths of different approaches, but it can also mean increased complexity. In short, it is easy for researchers to obtain the SCM through any type of approach. Therefore, they can choose different methods according to the modeling need, characteristics of the dataset, and personal preferences. 

\section{Semidefinite Task}
This section focuses on the construction of a semidefinite structural  causal model (SSCM) consisting of binary-variable dataset (BVD) under the semidefinite task. In section \emph{A}, We define semidefinite task. Then, we separately define and instantiate the BVD and the SSCM in section \emph{B} and \emph{C}. In addition, we propose some causal discovery methods for SSCM in sections \emph{D} and \emph{E} respectively, summarizing datasets used for SSCM at the same time. Finally, we summarize these approaches in section \emph{F}.

\subsection{Definition of Semidefinite Task}
According to the definition of causality, causal discovery makes sense in at least three variables in general. But in some bivariate data only have endogenous variables $V$ and $W$, exploring the potential causal relationship in these samples also appeared. Under pattern recognition is difficult to significantly improve, the thinking of false-correlation gradually increases. One of the most widely studied ways related with false-correlation is fine-grained recognition\cite{deng2013fine,duan2012discovering,krause2014learning}, because we can extract multiple highly similar local regions' points as new multiple variables from one sample, further to predict the label variable according to their relationships. In light of this, we define semidefinite task below to discover the causality between two endogenous variables. Because so few variables are known, such tasks still need to consider some exogenous variables $Z_1,Z_2,\dots,Z_t$ that are not generally recognized to be related to $V$ and $W$ or difficult to sample directly.

\begin{figure}
  \includegraphics[width=0.95\linewidth]{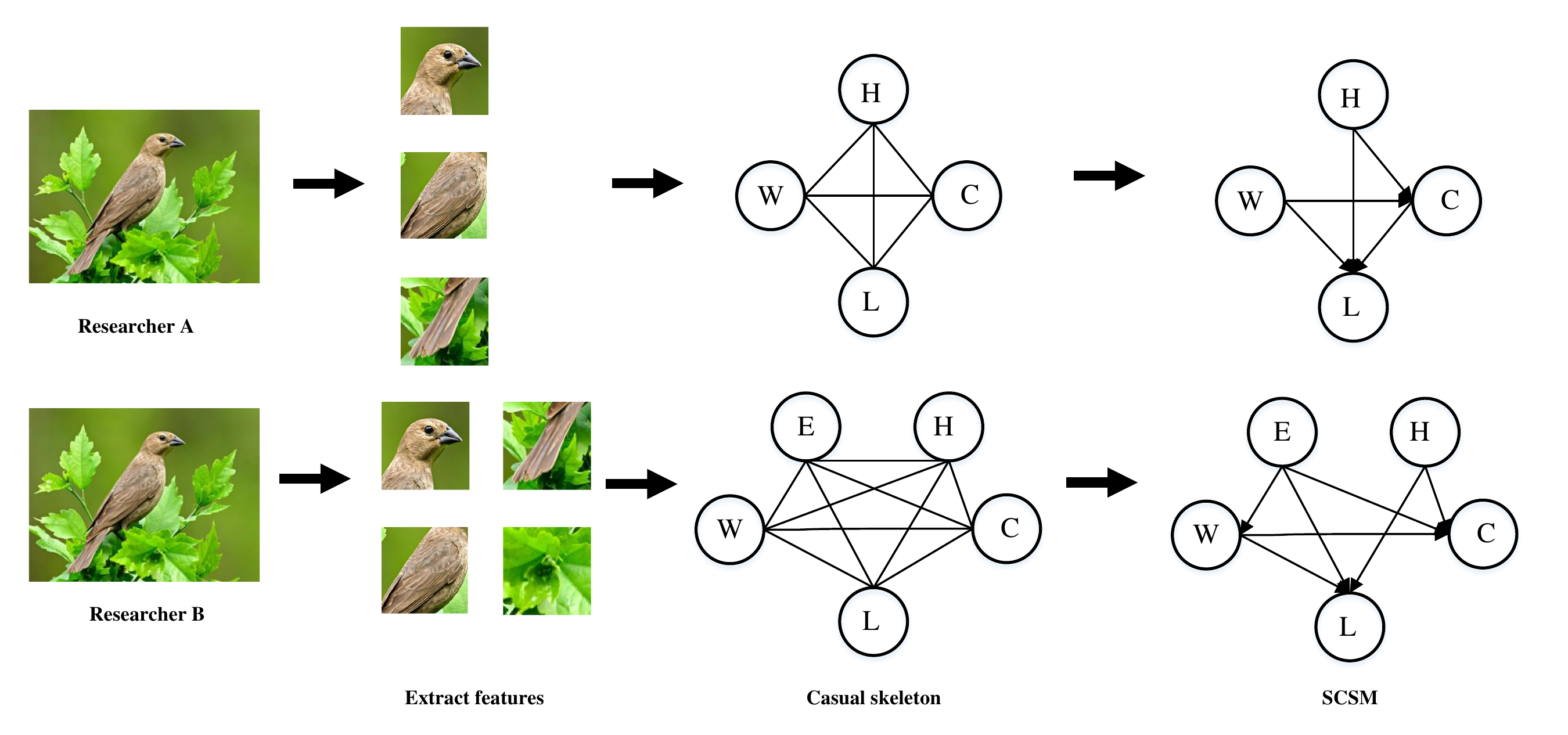}
  \caption{An Example of Semidefinite Task to illustrate its characteristics. Different researchers A and B may extract different points from the same image, so that the causal skeleton and the SSCM are different in the end.}
  \label{semibird}
\end{figure}

\textbf{Definition 4 (Semidefinite Task)} Explore the causality between $V$ and $W$, under the influence of the variables $X_1,X_2,\dots,X_s$ and $Z_1,Z_2,\dots,Z_t$, where $s = 1,2,\dots,p$ and $t = 1,2,\dots,q$ are unknown but finite.

Semidefinite task usually studied the influence of unknown but finite variables between two variables when $s = 0$. Since there is no additional information to know the number of exogenous variables $Z_t$, it is up to the researcher to determine the number according to their prior in the task. In addition, $V$, $W$ and $X_i$ are not equivalent here, and this task focus to obtain the direct or indirect causal relationship between variable $V$ and its label $W$. Semidefinite task applies to binary-variable datasets and aims to construct semidefinite structural  causal model in the end, the dataset and SSCM we will define below.

Let us show a bird dataset as an example. The CUB200-2011\cite{wah2011caltech} dataset contains 11,788 images of 200 bird species, with each sample containing two variables: the image ($V$) and the bird species ($W$). In order to distinguish the species of bird in the image, different researchers will divide the picture into multiple variables according to different types to derive some variables like $Z_1,Z_2,\dots,Z_t$. So that they can obtain the influence caused by exogenous variables. As shown in Figure~\ref{semibird}, researchers consider that the relationship between several parts might be helpful to classify. Such as researcher A thought head, wings and tail are important factors, because some birds may have a special tail or larger wings, as well as a different shape of the head, while researcher B may add additional environmental factors. Therefore, they can obtain the initial causal skeleton in theory through variables they extracted from the different parts of the image, and further construct SCMs using causal discovery approaches.

All in all, we summarize the feature of semidefinite task and its difference from definite task: (1) semidefinite task is applicable in BDV, but definite task is applied to MVD; (2) semidefinite need to construct causal skeleton according to researches' prior, but definite task van construct it directly; (3) semidefinite task construct multiple SCMs in the end, but definite task constructs only one. Therefore, the definite part of semidefinite task is the binary variables $V$ and $W$, they are exogenous and given in advance. In contrast, the endogenous variables $Z_1,Z_2,\dots,Z_t$ are unknown in a task, they are varying with each individual, so the unknown of endogenous variables is the indefinite part of semidefinite task.

\subsection{Binary-Variable Dataset}
Our defined binary-variable dataset with two exogenous variables, usually called data and ground truth, applide in pattern recognition and classification tasks in machine learning generally. According to the characteristic of semidefinite task, since $V$, $W$ and $X_i$ are usually not equivalent  when $s = 0$ or $s \neq 0$, we can give the definition of BVD.

\textbf{Definition 5 (Binary-Variable Dataset)}
A set is in form of $\left\{(V,W)=(v^{(j)} ,w^{(j)}), j=1,2,\dots,k\right\}$ is binary-variable dataset (BVD) in narrow sense, and it can also in form of multi-variable dataset $\{(V,W,X_1,X_2,…X_s)=(v^{(j)},w^{(j)},x_1^{(j)},x_2^{(j)},…,x_s^{(j)}), j=1,2,…,k\} $ in broad sense, where $(V,W)=(v^{(j)},w^{(j)}) $ are the main variables of BDV, and the remaining variables are the secondary data.

The data in BDV is in form of multidimensional space-time sequence, usually as pictures, text, audio, etc. As above mentioned, we can summarize some characteristics of BVD: (1) There are two given variables, the data contain many information and the ground truth; (2) According to researchers' prior knowledge and modeling needs, the dataset needs to be processed in some way of extracting points or deriving variables before using.

\subsection{Semidefinite Structural Causal Model}\label{sec3.3}
In definite task, all variables are given in MVD, so we can construct a causal skeleton directly. But in the semidefinite task, since the known binary variables in BDV cannot be directly modeled, we need to extract other variables from samples to model the causal skeleton, so the causal discovery step proposed in the DSCM is no longer applicable here, and we need to define new SCM applicable to other methods. Therefore, we can in line with BDV to define the SSCM.

\begin{table*}
  \caption{Examples of BDV in text domain.}
  \footnotesize
  \centering
  \begin{tabular}{|c|c|c|c|}
    \hline
    \textbf{Dataset}&\textbf{Training Set} &\textbf{Test Set}& \textbf{Introduction}\\
    \hline
    Amazon reviews\cite{mudambi2010research} & 346,686,770& - &\thead{Amazon.com product reviews, goods related to books, \\electronics, clothing and other 26 categories.}\\
    \hline
    \thead{Yelp reviews\cite{luca2016reviews}} &130,000&10,000&\thead{Merchant review site user review data in Yelp, \\including restaurants, shopping centers, hotels, \\tourism and other areas of merchants around the world.}\\
    \hline
    \thead{Yelp polarity reviews\cite{luca2016reviews}} & 280,000&19,000&\thead{Based on the previous dataset, binary labels \\such as positive or negative are included.}\\
    \hline
    IMDB\cite{oghina2012predicting} & 25,000&25,000&IMDB movie review data\\
    \hline
    CNC\cite{tan2022causal} &3,559&- &Social News Data Corpus\\
    \hline
    OntoNotes$5$\cite{dror2017replicability} & 6,672 $\sim$ 34,492&280 $\sim$ 2,327&\thead{Large news corpus, including 7 types.}\\
    \hline
    AG's news corpus\cite{zhang2015character} & 30,000&900&Classified coverage from over 2,000 news sources.\\
    \hline
    Sogou news corpus\cite{zhang2015character} & 90,000&2,000&\thead{A combination of SogouCA and SogouCS news \\corpus, containing various topic channels.}\\
    \hline
    CFPB compliant\cite{tan2022causal} & 555,957&- &\thead{U.S. Consumer Financial Protection \\Bureau (CFPB) Consumer Complaint Data.}\\
    \hline
    Gender Citation Gap\cite{maliniak2013gender} & 3,201&- &\thead{Data from the TRIP database obtained \\from the IR literature since 1980.}\\
    \hline
    Weiboscope\cite{fu2013assessing} & 4,685&-&\thead{Real-time posting data of \\Weibo social media users.}\\
    \hline
    SNIPS\cite{larson2019evaluation} & 13,084&700&\thead{The personal voice assistant collects \\data containing 7 domains}\\
    \hline
    CAIS\cite{liu2019cm} & 7,995&1,012&\thead{Voice collection by artificial \\intelligence speakers on 11 topics}\\
    \hline
    CrisistextLine\cite{gould2022crisis} & 10,118&55,473&Session data of counselor and mentee.\\
    \hline
    JDDC\cite{chen2019jddc} &1,024,196 & - &\thead{JD customer service staff on \\the topic of after-sales conversations}\\
    \hline
  \end{tabular}
  
  \label{tabtextbdv}
\end{table*}

\textbf{Definition 6 (Semidefinite Structural  Causal Model)} 
Take a definite $q$ firstly and under the priori assumptions, get the distribution of $Z_1,Z_2,…,Z_t  (t=1,2,…q)$ or some kind of its sampling. let $\mathcal{V} =(V,W,X_1,X_2,…X_s,Z_1,Z_2,…,Z_t )$, then the directed acyclic graph set $\mathcal{G} =(\mathcal{G} _1,\mathcal{G} _2,…,\mathcal{G} _m)$ constructed form $\mathcal{V} $ is a semidefinite structured causal model (SSCM), where $\mathcal{G} _j=(\mathcal{V} _j,\mathcal{E} _j)$ and the edge  $\mathcal{E}_j$ in the graph represents the causal relationship between variables.

As mentioned previously, $s$ is usually $0$ in semidefinite task, so USCM usually only needs to consider two exogenous variables $V$ and $W$, and the exogenous variables $Z_1,Z_2,…,Z_t$ need to be confirmed in quantity $q$. Moreover, only the one-way relation of $V$ and $W$ is considered, but they are equivalent in causal relationships. Due to $q$ represents the maximum of exogenous variables that can be derived from a BVD, it depends on the sample size of the dataset, so it can also be named as the sampleable critical point of a BVD. Moreover, although different researcher may divide and extract different $Z_1,Z_2,…,Z_t$ depend on their prior, further to construct different causal skeleton, there is still a limit to $m$, which represents the number of causal skeletons a BVD can construct. Because it depends on the information content, m should be in the range of $1 \leq  n \leq  \frac{IIN}{FBIN}$ , where $IIN$ is the information content of input data and $FBIN$ is the information content of each separated feature block.

In addition, due to different researcher has different priors and task expects, the final SSCM they constructed is diverse. And it is difficult to analyze the quality of different SSCM directly from the result indicators, because the result indicators are depending on many aspects, such as the computational power of the model, the upper limit of the dataset itself and so on. Therefore, we propose some metrics to describe some visualization results indirectly related to SSCM performance, where $q$ and $m$ mentioned above can be used to describe it: (1) The smaller $q$ means higher sampling, and there is a stronger hypothesis of SSCM; The larger $q$ or it's close to the critical point of sampleable , means lower sampling and SSCM with relatively weaker hypothesis. (2) The lager $m$ means the more point blocks are divided and extracted from data, so the coincidence rate of nodes in SSCM will be higher, and it's easier for SSCM to overfit; In contrast, the smaller $m$ means the number of SCM nodes is smaller, it will lead to have much confounders we didn't take that into account.

\begin{figure}
  \includegraphics[width=0.95\linewidth]{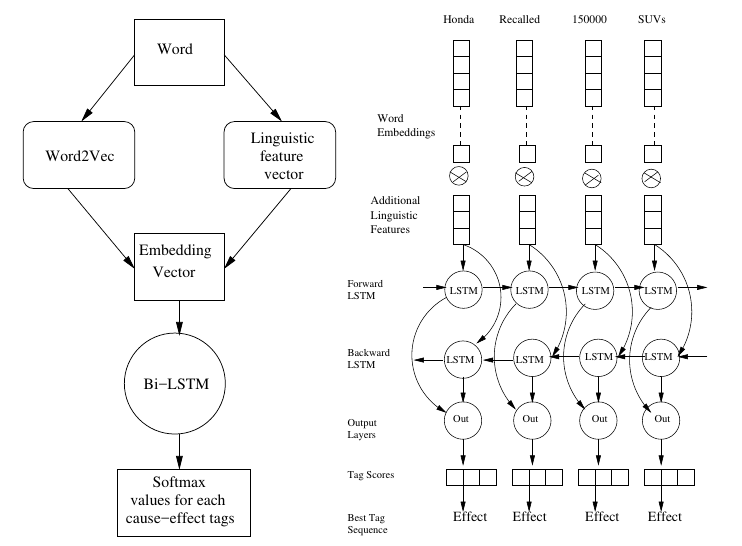}
  \caption{Overview of the bidirectional LSTM architecture for Cause-Effect relation extraction.\protect\footnotemark[1] }
  \label{bcere}
\end{figure}
\footnotetext[1]{The Figure~\ref{bcere} is driven from ~\cite{dasgupta2018automatic}.}

All in all, with SSCM, we can obtain the relationship between causal variables in the semidefinite task and facilitate the results of the pattern recognition task. We summarize the characteristics of SSCM as follows: (1) In SSCM, the number of all variables are definite before modeling, containing the endogenous variables and the exogenous variable depended on researchers' prior; (2) The final SSCM constructed is indefinite, its form of existence is diverse because of the indefinite content of the endogenous causal variables.

Another significant aspect of SSCM is how to construct it. As we discussed above, the data in BDV is in form of multidimensional space-time sequence, and a BDV will not construct a unique SSCM in the end, so we cannot use the causal discovery methods we proposed in DSCM for whole dataset to construct SSCM. In the next two sections, we summarize some studies for semidefinite task in image and text domains.

\subsection{Examples in Text Domains}
Causality in the text domain exists in various forms, it is commonly expressed as explicit and implicit causality, as well as marked and unmarked causality. Marked causality refers to the presence of linguistic signals of causality\cite{stukker2012subjectivity,ali2021causality}, such as the presence of causal markers like "because" and "so". Explicit causality means both cause and effect are expressed in the sentence, and implicit causality may lack the expression of the cause or effect. Many explorations of causality and models have emerged in the text domain, and these fall under the semidefinite task we defined. Automatic extraction of textual causality\cite{asghar2016automatic,dasgupta2018automatic} is mainly through three different approaches: linguistic rule-based\cite{ben2011automatic}, supervised\cite{sharp2016creating}, and unsupervised machine learning methods. The initial extraction methods focus on extracting causal pairs of individual words, and this linguistic rule approach relies heavily on domain and linguistic knowledge, making it difficult to scale up. Due to the presence of implicit causality, nested causal expressions and so on, traditional machine learning methods have emerged, but machine learning methods also rely heavily on feature engineering. In light of these problems, we propose two effective deep learning-based methods here, both of them are not limited to extracting causal relations of individual words and also avoid the above problems.

\textbf{BCERE (Bidirectional LSTM Architecture for Cause-Effect Relation Extraction):} A linguistic information-based deep neural network architecture\cite{dasgupta2018automatic} for automatic cause-effect relationship extraction from text documents, using word-level embeddings and other linguistic features to detect causal events and their effects.

The architecture applies a bidirectional LSTM to detect causal instances from sentences, as shown in Figure~\ref{bcere} The problem of insufficient training data is also addressed by using additional linguistic feature embeddings instead of regular word embeddings. By using this deep language-based architecture, phrases are effectively extracted as events, avoiding the erroneous results caused by extracting individual words as events like many existing tasks, as well as complex feature engineering tasks.

\textbf{CLCE (Conceptual-Level causality extraction):} This method\cite{mirza2016catena} aims to identify causal relationships between concepts, also named between linguistic variables, and represent the output in the form of a causal Bayesian network. This idea breaks the existing methods that mostly can only extract low-level causality between single events, and allows the machine to extract causality between disjoint events.

The method is divided into three subtasks: extracting linguistic variables and values, identifying causality between extracted variables, and creating conditional probability. We first need to extract the linguistic variables and values from the corpus, for example, the linguistic variable can be " Age ", and its value can be " Young ", "Old" and other specific descriptions. Then according to the causal markers in PDTB and the set of verbs in AltLexes, the authors create a causal database, and further define the causal relationship between two concepts or linguistic variables. Finally, the normalized pointwise mutual information scores were used to calculate the probability distributions of the linguistic variables and represent them in a causal Bayesian network.

In addition, we summarize some relevant datasets involving text for causal discovery and inference below, as shown in Table~\ref{tabtextbdv}. Amazon reviews\cite{mudambi2010research} and Yelp reviews\cite{luca2016reviews} are two common large user review corpora, and they can be used to build SSCM to identify review propensity or product category through reviews. IMDB\cite{oghina2012predicting} movie review data can similarly be used to build SSCM to identify the category of the film and the emotional orientation of the reviews. CNC\cite{tan2022causal}, IAC\cite{abbott2016internet}, OntoNotes 5\cite{dror2017replicability}, AG's news corpus\cite{zhang2015character}, and Sogou news corpus\cite{zhang2015character} are several multi-category news corpora with different sample sizes, they can be used to identify causal relationships in news statements or to build SSCM to identify news categories and sentiment tendencies. CFPB complaint\cite{tan2022causal} is consumer complaint data from the U.S. Consumer Financial Protection Bureau (CFPB) with the category label of whether the complaint was resolved promptly, it can be used for the semidefinite task of identifying features of complaints resolved. Gender Citation Gap\cite{maliniak2013gender} is literature data from TRIP with the ground truth of literature citation, previous studies have used this dataset to study the influence of author gender on citations. Considering the article topics will also affect citations, we can extract topics from the text as new variables, and construct SSCM by combining author gender, article citations, etc. Weiboscope\cite{fu2013assessing} dataset is a real-time posting data of Weibo social media users, it can be used to investigate the causal effect of censorship experience on subsequent censorship and posting rate of Chinese social media users. SNIPS\cite{larson2019evaluation} and CAIS\cite{liu2019cm} are all discourses on different topics collected by AI voice assistants, they can be used for causal discovery between utterances or constructing SSCM for classification according to features in the text. CrisistextLine\cite{gould2022crisis} is a corpus of conversations between counselors and counseling clients, it can also be used by some researchers to calculate and evaluate the causal effect between assignment tendency and counseling effectiveness using text as a mediating variable. JDDC\cite{chen2019jddc} is a conversation data from Jingdong customer service staff about after-sales topics, it can be used to analyze the reasons for after-sales consultation of not similar products.

\begin{figure}
  \includegraphics[width=0.95\linewidth]{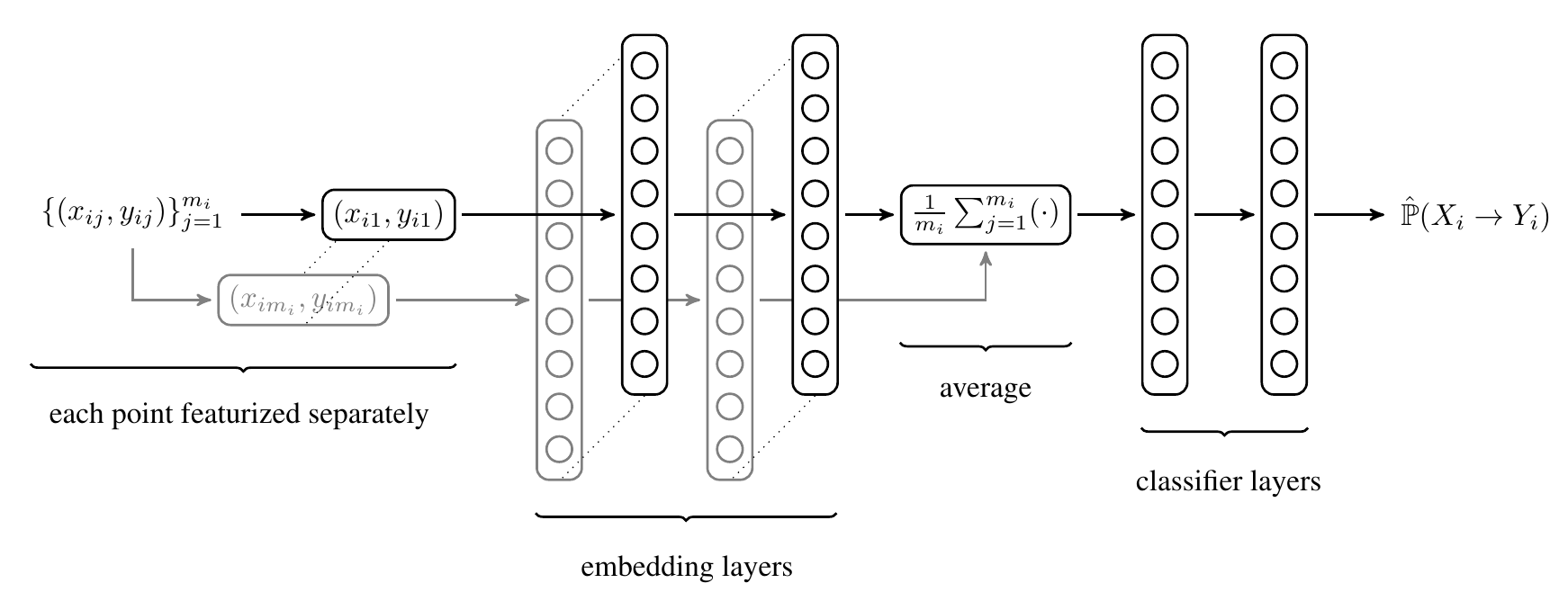}
  \caption{The architecture of Neural Causation Coefficient (NCC)\protect\footnotemark[2]. NCC maps the data points $\{(x_{ij},y_{ij})\}_{(j=1)}^m$ of the two variables to the multi-dimensional feature hidden layer, and finally brings the averaged results into the binary classifier to obtain the causal direction between the two variables. }
  \label{ncc}
\end{figure}
\footnotetext[2]{The Figure~\ref{ncc} is driven from ~\cite{lopez2017discovering}.}

\subsection{Examples in Image Domain}
Turning now to other domains, semidefinite tasks also exist in the field of vision, and they are mainly carried out according to image datasets. Image datasets usually contain only two variables, image and ground truth, so the researcher is required to extract other variables, also called causal features, from the image features to perform causal discovery. This type of semidefinite task often identifies the deeper features of the image and thus better contributes to the results of pattern recognition. In this paper, we summarize three kinds of semidefinite tasks in the image domain, they are studied from different angles according to the causal signals present in the image.

\textbf{NCC (The neural causation coefficient method):} A method\cite{lopez2017discovering} to reveal causal relationships between pairs of entities in an image, it confirms that there are observable signals in the image. The signals can explain causal relationships, also mean the higher-order attributes of the image dataset contain informations about the causal distribution of the objects. The method generalizes the features in images into object features, context features, causal features, and anti-causal features. The object features are those mostly activated inside the bounding box of the object of interest, and the context features are mostly activated outside the bounding box of the object of interest. Independently and in parallel, causal features are those that cause the presence of the object in the scene, whereas anti-causal features are caused by the presence of the object in the scene. Considering this, the method assumes there is a clear statistical dependency between object features and anti-causal features, while statistical dependence between context and causal features is absent or much weaker. This hypothesis is a support for the presence of causal signals in the image, and is also proven to exist.

\begin{figure}
  \includegraphics[width=0.95\linewidth]{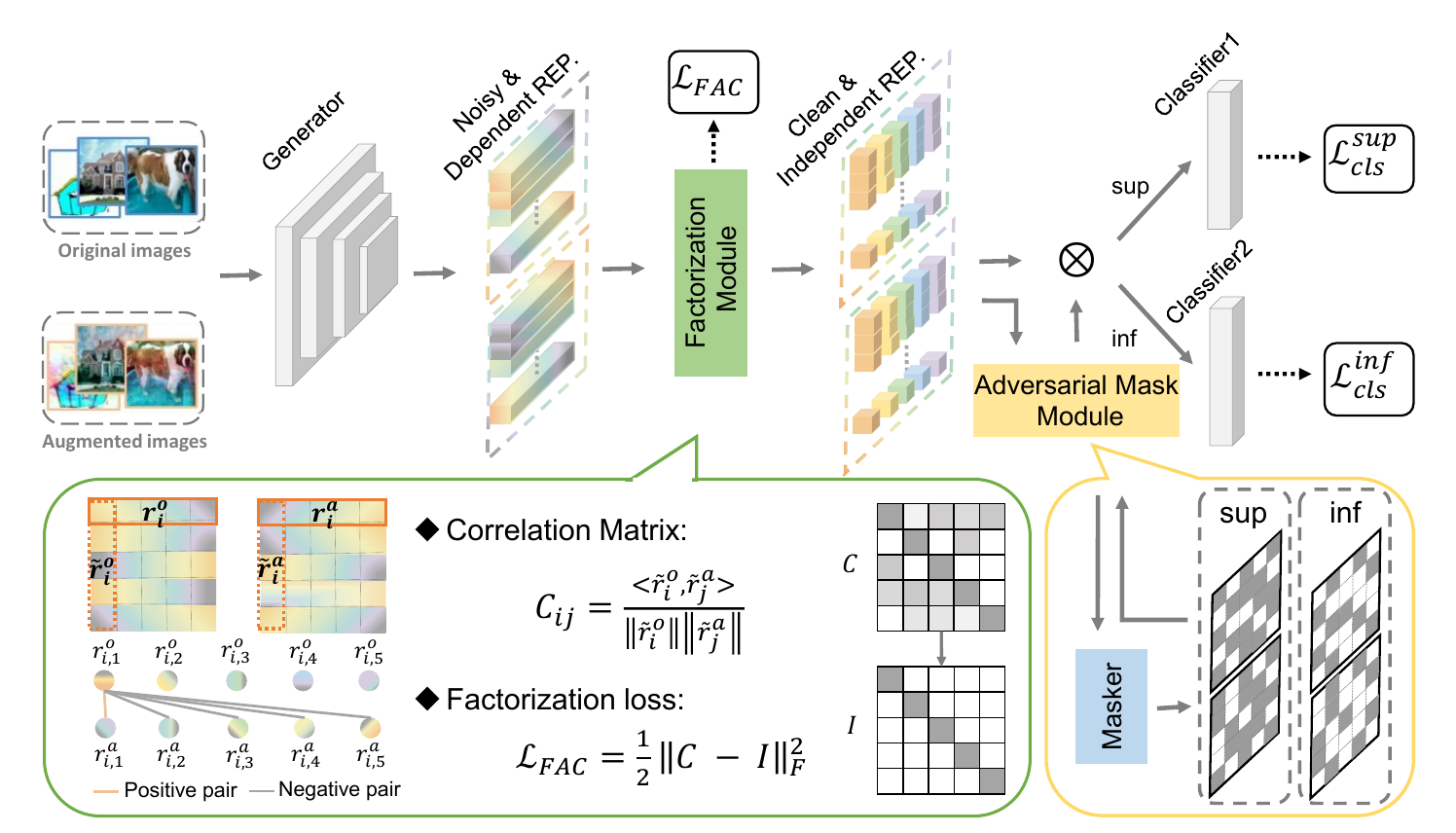}
  \caption{The framework of CIRL.\protect\footnotemark[3] It consists of three modules: causal intervention module, causal factorization module and adversarial mask module. The input is an image, and the output is the causal direction of the causal factor. }
  \label{cirl}
\end{figure}
\footnotetext[3]{The Figure~\ref{cirl} is driven from ~\cite{lv2022causality}.}

The NCC method is similar to the algorithms based on functional causal models to determine the causal direction mentioned in the previous section, it considers the independence between the cause and the mechanism (ICM) only satisfied in the correct causal direction $X \rightarrow  Y$, and it is not satisfied under the wrong causal direction $Y \rightarrow  X$. In this way, the causal direction between the main variables of interest will leave detectable features in their joint distribution. Therefore, NCC introduces neural networks to learn such detectable causal features. For ordinary binary data points $\{(x_{ij},y_{ij})\}_{(j=1)}^m$, it can be directly brought into NCC; for binary data sets containing only image and the ground truth, NCC will be applied to the scatter plot $\{(f_{jl},y_{jk})\}_{(j=1)}^m$ where $k$ corresponds to the number of category labels, and $l$ corresponds to the number of features which is also the number of hidden layer units of the neural network. Then the features are mapped to multidimensional for each point, and finally to a binary classifier to obtain whether the two variables belong to a causal or non-causal relationship.

\begin{figure}
  \includegraphics[width=0.95\linewidth]{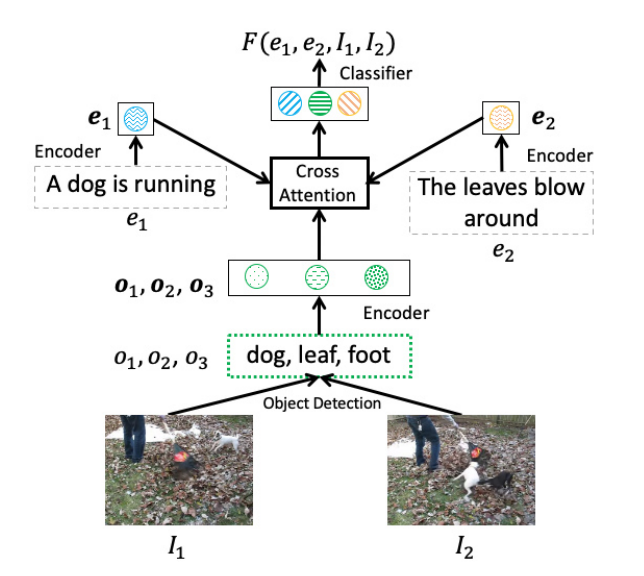}
  \caption{Demonstration of the proposed model. \protect\footnotemark[4] The method first extracts the events in the image, and then identifies the causal relationship between the events. }
  \label{vcc}
\end{figure}
\footnotetext[4]{The Figure~\ref{vcc} is driven from ~\cite{zhang2021learning}.}

The proposed NCC effectively distinguishes causal and anti-causal features, possessing a relatively low computational complexity. Its complexity in forward annotating is linearly related to the sample size, whereas the computational complexity of the sota kernel-based additive noise models is the cube of the sample size. And NCC can be trained using a mixture of different causal and anti-causal generative models for a wide range of applications. Its differentiable activation function can also embed NCC into larger neural architectures. In light of this, methods such as HDCC\cite{sisi2020finding}, which extends causal inference to more than two variables, have also been derived.

%加入表格
\begin{table*}
  \caption{Examples of BDV in image domain.}
  \footnotesize
  \centering
  \begin{tabular}{|c|c|c|c|}
    \hline
    \textbf{Dataset}&\textbf{Training Set} & \textbf{Test Set} &\textbf{Introduction}\\
    \hline
    CIFAR-10 & 50,000 & 10,000 &\thead{Image dataset with 10 types\\ of pervasive objects.}\\
    \hline
    \thead{CIFAR-100} & 50,000 & 10,000 &\thead{Image recognition dataset with 100 classes, \\each respectively containing fine-grained \\and coarse-grained labels representing the image.}\\
    \hline
    ImageNet & 14,197,122 & - &\thead{More than 20,000 categories \\of image recognition data}\\
    \hline
    ObjectNet\cite{barbu2019objectnet} & 50,000 & 10,000 &\thead{Over 300 categories of image recognition \\data, objects are shown in different \\camera angles in a cluttered room.}\\
    \hline
    PPB\cite{buolamwini2018gender} &1,270 & - &\thead{A facial dataset constructed from photographs \\of congressman from six different countries, which \\can be categorized by gender or skin color.}\\
    \hline
    YFCC-100M Flickr\cite{karkkainen2021fairface} &108,501& - & Facial recognition data\\
    \hline
    Casual Conversations\cite{hazirbas2021towards} & 45,000 & - &\thead{Facial recognition dataset that can include age, \\gender, and skin color as classification tags.}\\
    \hline
    CelebA\cite{liu2018large} & 202,599 & - & Facial feature recognition data\\
    \hline
    Aff-wild2\cite{kollias2018aff} & 260 & - &\thead{The video data of facial recognition downloaded \\from Youtube showed a wide variation of \\subjects in terms of posture, age, lighting \\conditions, race and occupation.}\\
    \hline
    WEBEmo\cite{panda2018contemplating} & 26,800,000 & - & \thead{Facial emotion recognition data with \\binary emotion category labels for classification.}\\
    \hline
    Market1501\cite{zheng2015person} & 12,936 & 19,732 &\thead{Pedestrian re-identification dataset.}\\
    \hline
    CUB200-2011\cite{wah2011caltech} & 5,994 & 5,794 &\thead{Includes identification data for \\200 different bird species.}\\
    \hline
    Stanford Cars\cite{fu2017look} & 8,144 & 8,041 & Car model identification dataset.\\
    \hline
    Veri-776\cite{lou2019veri} & 50,000 & - &Vehicle re-identification data.\\
    \hline
    Cityscapes\cite{hu2020probabilistic} & 2,975 & 1,525 & Street scene recognition data from 50 different cities.\\
    \hline
  \end{tabular}
  \label{tabimagebdv}
\end{table*}

\textbf{CIRL (Causality Inspired Representation Learning for Domain Generalization):} A representation learning method\cite{lv2022causality} according to causal mechanisms for solving the domain generalization problem prevalent in image classification scenarios in CV, and is an application of causal discovery to domain generalization. It solves the problem of modeling the relationship between data and ground truths with statistical models that cannot be generalized when the distribution changes. The framework of CIR as shown in Figure~\ref{cirl}.

The method assumes each input is composed of causal and non-causal factors, and only the former causes the classification judgments. By the nature of the Common Cause Principle and the Independent Causal Mechanisms Principle, causal factors are separated from non-causal factors, are joint independent and causally sufficient for the classification task $X \rightarrow  Y$ are causally sufficient. The method aims to extract causal factors from the input and then reconstruct the invariant causal mechanisms. The causal learning representation algorithm consists of three modules: causal intervention module, causal factorization module and adversarial mask module. The causal intervention module generates enhanced images, and intervene on the non-causal factors. Both the original and the enhanced image representations are sent to the causal factorization module, this model imposes a decomposition loss to force the representation to be separated and jointly independent from the non-causal factors. Finally, the adversarial mask module conducts an adversary between the generator and a masker, rendering the representations to be causally sufficient for classification.

The method effectively solves the problem of statistical models in domain generalization cannot be generalized, reconstructs the causal factors, and uncovers the intrinsic causal mechanisms, so that the causal representations learned by the CIRL framework can model the causal factors based on the ideal properties we emphasize.

\textbf{VCC (Vision-Contextual Causal Model):} A method\cite{zhang2021learning} for learning contextual causal relationships from visual signals, and explores the possibility of learning causal knowledge from time-consecutive images. The method points out some causal relationships exist only in specific contexts, so the method acquires causal knowledge from time-consecutive frames cropped from the video, and find all possible causal relationships between each image pair $P$ contains two images $I_1$ and $I_2$, it requires to identify all the events in the images first before to identify the causal relationships between the events.DThe emonstration of the proposed model as shown in Figure~\ref{vcc}.

The VCC model consists of three main components: event encoding, visual context encoding, and cross attention. First, in the event encoding module, events represented in natural language are transformed into vectors and encoded using the BERT\cite{devlin2018bert} model. Then the object detection module\cite{xu2017scene} is used to detect objects from images and use all extracted objects to represent the visual context. The cross-attention module is used to select the important context objects to reduce the effect of noise. Finally, the causality score can be predicted to determine the causality of the context.

Here, we list some common BDVs in the image domain, as shown in Table~\ref{tabimagebdv}. The first major categories such as CIFAR-10, CIFAR-100, and ImageNet are commonly used pervasive object recognition datasets, all of them contain only two variables, images, and their category labels, with the difference in sample size and label classes. ObjectNet\cite{barbu2019objectnet} is an image recognition data containing more than 300 categories, objects in the picture are displayed in a messy room with different shooting angles, they can be used to identify the causal relationship between the main object in the picture and the surrounding scene. The second major category is the facial recognition dataset, they can be used for the important downstream task of face recognition. PPB\cite{buolamwini2018gender} is a facial dataset constructed by collecting photos of members of parliaments in six different countries, and its collection process takes into account the possible biases that can occur when building structural  causal models with gender or skin color as category labels. The YFCC-100M Flickr\cite{karkkainen2021fairface} facial recognition dataset was originally collected to reduce bias in facial recognition for Caucasians, gender, skin color, and age can be selected as category labels for recognition, and likewise the Casual Conversations\cite{hazirbas2021towards}, CelebA\cite{liu2018large} facial recognition dataset. Aff-wild2\cite{kollias2018aff} downloaded video data for facial recognition from Youtube, subjects had posture, age, lighting conditions, race, and occupation very different, and these features can be considered as dependent variables. WEBEmo\cite{panda2018contemplating} is face recognition data suitable for binary emotion classification, it is widely used for facial emotion recognition tasks, and similarly for several datasets such as CK+, MMI, Oulu-CASIA, Deep Sentiment, and emotion-6. The third category of image data belongs to a large class and is usually used for individual or model recognition tasks within the same superclass. Market1501\cite{zheng2015person} pedestrian re-identification data can be refined to local area features for fine-grained recognition of different pedestrians, and similarly, the Veri-776\cite{lou2019veri} vehicle re-identification dataset. CUB200-2011\cite{wah2011caltech} contains image data of 200 different bird species, FGVC-Aircraft\cite{rao2021counterfactual} contains image data of 100 different types of aircraft, and Cityscapes\cite{hu2020probabilistic} contains image data of 50 different street scenes recorded at different periods, all of them can be used to construct SCMS to explore causal signals in images using local features and data labels.

\subsection{summary}
In this section, we defined another common causal discovery task, and according to its characteristics defined the corresponding dataset and SCM, summarizing the existing research methods. As previously stated, we conducted a comparative study of the different methods, for instance, it is easier to divide feature blocks and extract prior consensus on image data than text data. Causal representations are intuitive and easy to represent in image data, however, we need word vectors to represent the data and then extract feature blocks from keywords according to the priori. This process may be affected by different writing styles, where features are less intuitive than in image data, so it is also different to avoid confounders we neglected and other biases. On the other hand, text data can intuitively convey some information to researchers to infer causality easily, such as text theme and emotion analysis, syntax analysis, etc. These advantages can help solve some difficulties and contradictions in image recognition. In addition, we propose some indicators to describe visual results indirectly related to SSCM performance. Although this cannot directly judge whether a SSCM is good or not, it can bring more interpretable content to subsequent experiments. All in all, the existence of SSCM in causal discovery is important and inevitable, it helps us better to learn sample representation and its context features through deep learning approaches.

\section{Undefinite Task}\label{undef}
This section focuses on the construction of undefinite structural  causal model (USCM) over infinite-variable dataset (IVD) in undefinite task. In section \emph{A}, we define the undefinite task in causal discovery. Then, we respectively define and instantiate the IVD and the USCM in sections \emph{B} and \emph{C}. Moreover, we show some IVD in the dialogue, audio and video domains. Finally, we discuss some similarities and differences between MVD, BVD and IVD in section \emph{D}.

\begin{figure}
  \includegraphics[width=0.95\linewidth]{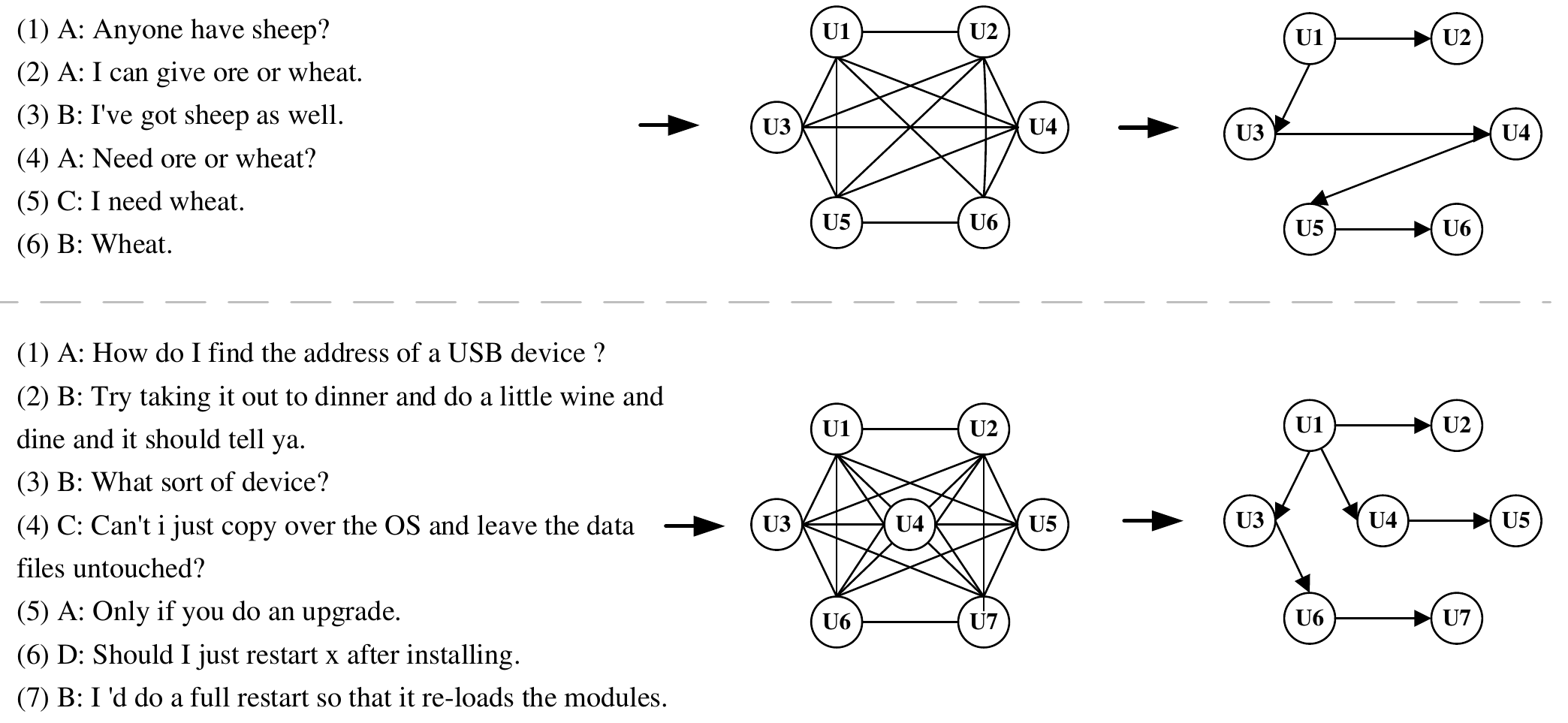}
  \caption{An Example of constructing USCM with IVD.}
  \label{utconversation}
\end{figure}

\subsection{Definition of Undefinite Task}
Besides definite and semidefinite tasks we defined previously, there is still another necessary task that has not been studied, corresponding to the causal discovery we proposed on sequence tasks. With the increasing demand of sequence tasks for sequence semantic understanding, the study of its causality is inevitable. Because the order of the sequence itself combines with semantic information to form a new order named causal order, that means the causal relationship in sequence can be obtained by coupling the order of the sequence with the semantic information, it can solve many problems. For instance, it accounts for the sequence fragments of the same relative position cannot be represented as the same node in the SCMs because of different semantic information, and the reason of study causal relationships in sequence tasks is because this is an enhanced sequence task essentially. Thus, we define undefinite task as follows:

\textbf{Definition 7 (Undefinite Task):} Explore the causality between $V$ and $W$, under the influence of the variables $X_1,X_2,…,X_s$ and $Z_1,Z_2,…,Z_t$, where $s=1,2,…,\infty $ and $t=1,2,…,\infty $ are unknown and undefinite.

Undefinite task is like the semidefinite task. Both need to consider the causal discovery of exogenous variables $Z_1,Z_2,…,Z_t$, and need to obtain the causal skeleton according to a priori. Due to the known variables $X_1,X_2,…,X_s$ do not provide the complete causal relationship, so additional variables still need to be introduced. But there are also significant differences between them: (1) Tasks apply to different datasets, undefinite task applied to IVD but semidefinite task applied to BVD. Variables V, W in undefinite task affected by unknown and possibly infinite variables are in more complex forms; (2) Tasks have a different prior basis to construct causal skeleton at present, undefinite task lacks recognized valid priors. This type of causal discovery task is suitable for infinite datasets and the final aim is to construct USCM, we will define them below.

\subsection{Infinite-Variable Dataset}
Under the above task definition, the ground truth of the data in the dataset used here is in the form of sequence, and the data is also in sequence form such as conversation sequences, audio sequences, video sequences, etc. So, there is each fragment in the sample corresponds to one ground truth, not only one ground truth for the whole sample. Each sequence sample in a dataset is almost different due to the length, meaning, and number of different sequence fragments consisting of it. Since $s$ and $t$ maybe $\infty $, we can define infinite-variable dataset contains $V,W,X_1,X_2,…X_\infty $ as follows:

%公式不连写
\textbf{Definition 8 (Infinite-Variable Dataset)} A multivariate set is in form of $\{(V,W,X_1,X_2,\dots,X_\infty )$ $=$ ($v^{(j)}$,$w^{(j)}$,$X_{j_1}$ $=$ $x_{j_1}^{(j)}$,$X_{j_2}$ $=$ $x_{j_2}^{(j)}$,…,$X_{j_l}$ $=$ $x_{j_l}^{(j)}$,$X_{else}$ $=$ $\varnothing $), $j$ = $1,2,…,k$,$l < \infty$, $j_l \in N _+$$\}$ is infinite-variable dataset (IVD), where $\varnothing $ is null set.

In general, the data and ground truth of IVD are both in multidimensional space-time sequences. Therefore, we should consider each labeled fragment as a variable to construct a causal skeleton, to avoid the bias in a sequence sample resulting from neglecting causal order affects the understanding of sequence context. Combing the existing form of data, the characteristics of IVD fall under two headings: (1) Each sample contains a different number of variables, and the length of each sample is infinite. For example, a conversation can be several seconds or several hours, and a piece of music can be two minutes or five minutes because there is no limitation to its length in practice. (2) Each variable in the dataset is different lie in its length and meaning, so even sequence fragments of the same position in different samples cannot be represented by the same variable. As discussed above, these are also the infinite property of IVD.  

Let us use a conversation sequence dataset\cite{li2020molweni} as an example for further explanation the content of each sequence fragment is different too. Molweni is an English corpus of multi-speaker conversations, where the first conversation is composed of 6 utterances from 3 speakers, and the second conversation is composed of 7 utterances of dialogue from 4 speakers. For each utterance $U_i$ in the conversation sequence, there is a ground truth $T_i$ for it, the keyword that the speaker wants to express. Two conversations are shown in Figure~\ref{utconversation}, the utterance \emph{(1)} in two conversations have different semantics because they are in different backgrounds, although they have the same order in conversation. If these variables are in the order of the data set itself, such as in time order, the fragments of the same relative position are represented  by the same relative time node, they can be considered as the same variable. But when these orders are mapped to causal orders, they cannot be considered as the same variable because they represent different semantics $U_{11}$ and $U_{21}$. Hence, we can't construct one causal skeleton for the two samples in an IVD, because each utterance is influenced by the speakers' style, the context and background the conversation takes place.

\subsection{Undefinite Structural Causal Model (USCM)}
Under the undefinite task, the undefinite structural causal model is established to explore the causality between different sequence fragment in IVD, and materialize the causal order between infinite variables in IVD according to some priors from researchers and causal discovery method, so its specific definition is as follows:

\textbf{Definition 9 (Undefinite Structural Causal Model):} With a definite prior, for each sample $\{(V,W,X_1,X_2,\dots,X_\infty )$ $=$ $(v^{(j)}$,$w^{(j)}$,$X_{j_1}$ $=$ $x_{j_1}^{(j)}$,$X_{j_2}$ = $x_{j_2}^{(j)}$,…, $X_{j_l}$ = $x_{j_l}^{(j)}$,$X_{else}$ = $\varnothing )$, $j = 1,2,…,k$, $l < \infty$, $j_l \in N _+$$\} $, let $V_j$ = ($V$,$W$,$X_{j_1}$,$X_{j_2}$,…,$X_{j_l}$,$Z_{j_1}$,$Z_{j_2}$,…,Z$_{j_t}$), then the set of directed acyclic graphs $\mathcal{G} = (\mathcal{G} _1,\mathcal{G} _2,…,\mathcal{G} _m )$ formed by them is undefinite structural  causal model (USCM), $\mathcal{G} _j = (V_j,\mathcal{E} _j )$ and the edge $\mathcal{E} _j$ represents causal relationship.

According to the definition of undefinite task and IVD, USCM has two main characteristics:
(1) The causal skeleton and SCMs constructed on each datapoint vary. As we above mentioned, different sequence samples in a dataset have different lengths, and each sequence fragment is also different. So the causal skeleton we construct for each sample with sequence fragments as variables is almost different, the same as for the final SCMs;
(2) The sparsity caused by sample space makes it impossible to sample variables in USCM. Regarding DSCM, the strong prior properties of MVD make all samples available for DSCM to construct the only SCM; Whereas in SSCM, due to the number of causal variables being definite, it can always find similar samples with the same label, although a dataset does not only construct one SSCM. But for USCM, the sparsity between variables directly leads to it being difficult to find a similar sample in IVD for one causal skeleton we mentioned previously. For example, in a conversation dataset like Molweni, the smallest labeled sequence in a conversation is an utterance ($U$). Considering the largest word list which can compose any $U$, each $U$ in the conversation is composed of $n$ words or phrases from this word list. The utterance $U$ can be treated as a sample in the word list set and $n$ is much smaller than the word number $N$ of the word list, so each utterance has significant sparsity. This property can also be mapped to sequence tasks like audio and video. Therefore, the $m$ of $\mathcal{G} _1,\mathcal{G} _2,…,\mathcal{G} _m$ in USCM exists $m \leq  s$ and $m$ is approximately $s$, where $s$ is the sample size of IVD. Hence, there are not enough samples to discover causal relationships for such data and USCM has non-samplability.

The process of constructing USCM is shown in Figure~\ref{utconversation}. For example in \emph{Molweni}, due to the existence of causal order in utterances, utterances at the same position in the different samples represent different variables in IVD and different nodes in SCM. In addition, different researchers may have different priori of causal order, which all lead to different SCM we finally obtained. Thus, the two samples construct two SCMs. Considering the specific case, when we compare two sequence samples with the same number of utterances, different utterances have different semantics. So that even if the number of sequence fragments is the same, the sequence samples cannot be represented by the same node in the SCM. Therefore, even if a sequence sample has the same number of sequence fragments, it cannot be represented by the same causal skeleton.

However, if a task is not to explore the causality between the sequence fragment, but to explore the causality between the whole sequence and some ground truth, the task turns undefinite into semidefinite. In this case, the causal relationship between sequence fragments still exists. If each fragment has a label, then we should divide the semidefinite task into two steps: the first step is to solve the undefinite task inside the model, and then according to the result solve the semidefinite. If there is no label for each segment and we still consider the relationship between segments, the final SCM we constructed is unqualified because the assumption is too weak to over the number of samplable information points we mentioned in section~\ref{sec3.3}.

Moreover, the non-samplability of USCM let us cannot obtain the definite causal skeleton, and there is also no suitable causal discovery method for constructing SCM. Undefinite task is still sequence task in essence, the key to solve the difficulties of this task is how to find a united prior rule to construct the different causal skeleton under the different variables, and to find the appropriate causal discovery methods instead of the method we mentioned previously in Sparse sample space. We'll discuss this in section~\ref{section5}. 

\subsection{Related Tasks and Datasets}
The sample data in the infinite-variable dataset is in the form of multidimensional space-time sequences, so dialogue, audio, and video data all conform to the form of such a sequence. In building a structural  causal model, each utterance of dialogue, each track or each lyric of a piece of music, each frame of a video, or the subtitle in that frame can be considered as nodes to construct a causal skeleton. In turn, by analyzing the context and data features, we can build a suitable structural  causal model. We have collated some of the available indeterminate-variable datasets and related tasks carried out on these data for researchers' reference.

\textbf{Conversation data.} One of the more novel types of UVD is conversation data, which includes task-based conversation, question-and-answer conversation, and small talk conversation. Task-based conversations are usually considered to be multi-round conversations, which are conversation service systems that perform a specific task in a specific context; question-and-answer conversations are mostly single-round conversations, that is one-question-and-answer forms, such as recognizing the intention of the questioner and selecting the answer from the knowledge base to return; and small talk conversations are not restricted in any way and are completely dominated and influenced by the interlocutor. Conversation datasets are mostly used for discourse parsing, and discourse parsing techniques have been successfully applied to tasks such as question and answer, text classification, sentiment classification and language modeling. The main approaches to parsing discourse in multi-person conversations are those based on manual feature construction and deep learning. However, these two approaches also suffer from the limitations of manually designed features and the limitations of deep learning models' generalization ability. Considering the specificity of the conversational corpus, graph structure-based approaches may be able to better analyze the inter-semantic relationships. We have compiled here some commonly used conversation data for different tasks, as shown in Table~\ref{tabconivd}.

\begin{table*}
  \caption{Examples of Conversation Datasets.\protect\footnotemark[5]}
  \footnotesize
  \centering
  \begin{tabular}{|c|c|c|c|}
    \hline
    \textbf{Dataset}&\textbf{Language} & \textbf{Examples of Tasks}& \textbf{Contents} \\
    \hline
    IEMOCAP\cite{busso2008iemocap} &English & \thead{ ERC\\NLU } &\thead{A conversation script for 10 male \\and female ctors during their emotionally \\binary interactions, containing 151 conversations, \\7433 utterances, 10 conversation roles, and 10 \\emotional labels.}\\
    \hline
    SEMAINE\cite{poria2019emotion} &English & \thead{ ERC } &\thead{A conversation between four fixed-image \\robots and a human being, labelled with four\\ emotional dimensions.} \\
    \hline
    MELD\cite{poria2018meld} &English & \thead{ ERC\\AR\\SR } &\thead{A selection of conversation clips from \\the movie Friends, including 1433\\ conversations and 13708 utterances, providing \\3 major categories of coarse-grained \\sentiment labels and 7 fine-grained sentiment \\labels.}\\
    \hline
    DailyDialog\cite{li2017dailydialog} & English & \thead{ ERC\\AR\\ECG } &\thead{A daily conversation dataset with 13,118 \\conversations, 102K conversational utterances, \\annotated with 7 emotions, 4 categories of \\conversational actions (DA) and 10 conversational \\themes.} \\
    \hline
    STC\cite{wang2020large} &Chinese & \thead{ ERC\\ODGG } &\thead{STC is an open domain Short-Text Conversation\\ dataset constructed from a corpus crawled from Weibo, \\containing 4.4 million conversations.} \\
    \hline
    DBCC\cite{wu2016sequential} & Chinese & \thead{ ERC\\ODGG } &\thead{A large open-domain conversation dataset \\from public social media, with mostly single-round \\conversations, including 1.4B conversations, \\and 3.0B utterances.} \\
    \hline
    WDC-Dialogue\cite{zhou2021eva} & Chinese & \thead{ ERC\\NLU } &\thead{A large open-domain conversation dataset \\from public social media, with mostly\\ single-round conversations, including 1.4B \\conversations, and 3.0B utterances.}\\
    \hline
    OpenSubtitles\cite{li2017dailydialog} & multilinguals & \thead{ ODGG } &\thead{Multilingual open domain conversation clips.}\\
    \hline
    Empathetic Dialogues\cite{welivita2021large} &English& \thead{ ERG } &\thead{Open domain conversation data, consisting \\of 25,000 conversations, provides 32 sentiment labels.}\\
    \hline
    PERSONAL-CHAT\cite{srivastava2022dictionary} &English & SR &\thead{Personalised conversation data,\\ including 10,981 conversations and 164,356 utterances, \\by 1,155 people involved in conversations.} \\
    \hline
    Ubuntu IRC Logs\cite{sinha2014investigating} &English& \thead{ CD\\DST } &\thead{From the Ubuntu IRC log dataset,\\ each chat room in the Ubuntu \\IRC is full of users discussing a \\variety of topics, mostly technical \\issues related to Ubuntu.} \\
    \hline
    Reddit Dataset\cite{turcan2019dreaddit} &English & DST &\thead{Consisting of posts and corresponding\\ replies in the Reddit forum, it contains a total \\of 4,500 conversations, \\120,000 utterances, and 35,000 speakers.}\\
    \hline
    STAC\cite{li2020molweni} &English& DST &\thead{The corpus is derived from player \\statements made during the online \\game \emph{The Setters of Catan} and contains 10,677 \\lines of conversation.} \\

    \hline
  \end{tabular}
  \label{tabconivd}
\end{table*}
\footnotetext[5]{In examples of tasks in Table~\ref{tabconivd}, ERC is Emotional Conversation Generation\cite{zhou2018emotional,shen2021directed}, NLU is Natural Language Understanding\cite{allen1995natural,liu2019multi}, AR is Action Recognition\cite{jhuang2013towards,poppe2010survey}, SR is Speaker Recognition\cite{campbell1997speaker,peskin2003using}, ECG is Emotional Conversation Generation\cite{sun2018emotional,peng2019topic}, ODGG is Open-Domain Dialogue Generation\cite{kann2022open}, ERG is Empathetic Response Generation\cite{majumder2020mime}, CD is Conversation Disentanglement\cite{li2020dialbert}, DST is Dialog State Tracking\cite{williams2014dialog,mrkvsic2015multi}. }

IEMOCAP\cite{busso2008iemocap} is an English dialogue corpus of actors' lines from TV scripts, collected during the emotional interaction between different pairs of actors, and is therefore often used for dialogue comprehension and dialogue emotion recognition tasks. MELD\cite{poria2018meld} is an English corpus of daily conversations with sentiment labels, action labels, and topic labels for sentiment recognition, action recognition and sentiment generation respectively. STC\cite{wang2020large} is an open-domain short-text conversation dataset constructed from a corpus crawled from Weibo, which can be used for open-domain conversation generation in addition to conversation sentiment recognition. Similar to the Douban Conversation Corpus\cite{wu2016sequential} from the Douban forum, and the WDC-Dialogue\cite{zhou2021eva} from public Empathetic Dialogues is also an open domain dialogue data, but it has been used to generate empathetic dialogues, based on which an empathy-oriented open domain dialogue model has been built. PERSONAL-CHAT\cite{srivastava2022dictionary} is personalized conversation data collected for different speakers, which contains at least five utterances for each speaker that highlight their personality, and can therefore be used for speaker identification tasks. As well as Ubuntu IRC Logs\cite{sinha2014investigating} from Ubuntu online chat rooms, which can be used for user profiling. Reddit Dataset\cite{turcan2019dreaddit} is multi-person conversation data from Reddit forums, and STAC \cite{li2020molweni} multi-person conversation data from online games, both of which can be used for multi-person conversation recognition tasks.

\textbf{Music data.} Similarly, music sequence data can be seen as Undefinite-Variable data for modeling purposes. Music sequences are a little more complex than contextual text-only conversation data, as they take into account not only the contextual relationship of the voice, but also the influence of other voices and tracks. Therefore, this paper compiles some common music data that contains only melodies but not lyrics.

\begin{table*}
  \caption{Examples of Audio Datasets.}
  \footnotesize
  \centering
  \begin{tabular}{|c|c|c|c|}
    \hline
    \textbf{Dataset}&\textbf{Sample Size} & \textbf{Contents}& \textbf{Tasks} \\
    \hline
    Bach Chorales dataset\cite{roman2019holistic} &300-400 &\thead{Bach's four-part congregational \\hymn dataset.}& \thead{Melodic generation, \\Harmonic modelling, \\Polyphonic generation,\\ Counterpoint generation}\\
    \hline
    Nottingham Dataset\cite{medeot2018structurenet} &1000 &\thead{Folk Song Folk Data.}& \thead{Style recognition, \\Melody generation,\\ Chord notation}\\
    \hline
    The conversation\cite{vercoe2001folk}&7000 &\thead{A melodic dataset of \\Irish folk music}& \thead{Melody generation}\\
    \hline
    The MAESTRO Dataset\cite{hawthorne2018enabling} &700-800 &\thead{Data set of the classical piano \\competition Piano-e-Competition}& \thead{Melody generation, \\Style recognition}\\
    \hline
    GiantMIDI-Piano\cite{kong2020giantmidi} &10,854 &\thead{The largest classical piano dataset.}& \thead{Music information retrieval, \\Automatic composition, \\Intelligent music composition}\\
    \hline
    MedleyDB\cite{bittner2014medleydb} & 196 &\thead{Polyphonic Pop Music Dataset}& \thead{Composition of polyphonic music,\\ Instrumental music generation, \\Polyphonic transcriptions}\\
    \hline
    Video Game Dataset\cite{politowski2020dataset} &31,810 &Music data in video games& \thead{Melody generation, \\Style recognition}\\
    \hline
  \end{tabular}
  \label{tabaudivd}
\end{table*}

The Bach Chorales dataset\cite{roman2019holistic} is a dataset of Bach's four-part congregational chants, containing at least four vocal parts: Alto, Tenor, Soprano, and Bass. Although this is a polyphonic dataset, each life part is monophonic. Therefore, if each vocal part is counted as a separate melody, the dataset is monophonic-monophonic. Similarly, there is the Nottingham Dataset\cite{medeot2018structurenet}, a more classic folk ballad dataset with individual chord markers that can be used for style recognition, melody generation, and chord annotation. The conversation\cite{vercoe2001folk} is a monophonic melody dataset of Irish folk music, uploaded by the user, which will have some repeated fragments and can be used for melody generation tasks. Data that is a little more complex than monophonic-monotone can be classified as a monophonic-polyphonic dataset. This type of data is most often found in music played on a piano, such as piano scores or MIDI where there is only one voice (instrument) on the piano, but the piano can be played with ten fingers, so up to ten notes can be pressed simultaneously. The MAESTRO Dataset\cite{hawthorne2018enabling} and GiantMIDI-Piano\cite{kong2020giantmidi} are two of the larger classical piano datasets, The MAESTRO Dataset is performance data from previous classical piano competitions and can be used for style recognition and melody generation. GiantMIDI-Piano is the latest collection of the world's largest classical piano dataset, which can be used for music information retrieval, automatic composition, and intelligent music creation. A little more sophisticated is the polyphonic dataset. Most polyphonic datasets contain piano guitars, which at once adds a strong polyphonic element. The content of common polyphonic-polyphonic datasets can be a whole orchestra of popular music or various repertoire, and the most complex is the score of a large orchestra, such as a symphony orchestra. A more difficult point with this kind of data than above is the need to distinguish between instruments/voices or to model them separately. It is possible that the amount of data available for certain voices, especially the backing voices, is quite small and insufficient to support modeling or training. medleyDB\cite{bittner2014medleydb} is an audio dataset of polyphonic music, that can be used for polyphonic music composition, instrumental music generation, and polyphonic transcriptions. video game Dataset\cite{politowski2020dataset} is a dataset of polyphonic music in video games that spans a wide range of time and game categories, and can be used for melody generation and style recognition.

\begin{table}
  \caption{Examples of Video Datasets.}
  \footnotesize
  \centering
  \begin{tabular}{|c|c|c|}
    \hline
    \textbf{Dataset}&\textbf{Sample Size} & \textbf{Contents} \\
    \hline
    YouTube - 8M\cite{abu2016youtube} &61,000 & \thead{Google, YouTube co-hosted \\a video tagging competition \\containing a large amount of \\video screen information, audio\\ information, and tagging \\information.}\\
    \hline
    UCF101\cite{soomro2012ucf101} &13,320 & \thead{An action recognition dataset\\ of real action videos with 101 \\action categories collected \\from YouTube.}\\
    \hline
    AVA\cite{murray2012ava}&6,849& \thead{A large collection of film \\and TV video clips from \\YouTube, with characters\\ involving professional actors \\of different nationalities.}\\
    \hline
    HMDB51\cite{carreira2017quo} &6,766 & \thead{A dataset of 51 categories \\involving facial movements and \\body movements collected from \\YouTube.}\\
    \hline
    Kinetics-600\cite{carreira2018short} &150,000 & \thead{Human-object interaction and \\human-human interaction video \\data collected from YouTube.}\\
    \hline
  \end{tabular}
  \label{tabvidivd}
\end{table}

\textbf{Video data.} Video data is also stored as sequences, which are also consistent with the characteristics of the proposed indeterminate-variable dataset. The video sequences are transformed into a frame-by-frame picture, that is the undefinite structural  causal model can be constructed in this unit. Video form datasets are commonly used for action recognition tasks, target segmentation tasks, target tracking tasks, and so on. In this paper, we have collated commonly used video datasets, as shown in Table ~\ref{tabvidivd}. YouTube - 8M\cite{abu2016youtube} is a competition video from YouTube, which contains nearly 4000 different categories of videos with large sample size and is commonly used for tasks such as scene detection, target tracking, and video classification. The multi-category video data collected from YouTube, with different motion patterns, atomic visual actions, and object interactions as classification criteria, have clear boundaries between the different classes, and the video data is highly stable and suitable for performing many different video recognition tasks.

\subsection{Comparing and Contrasting}
In this section, we defined a fresh causal discovery task, undefinite task, and according to its characteristics defined the corresponding dataset IVD and USCM, summarizing the existing different types of IVD. In addition, there are some features in IVD. Sequence data itself contains many causal relationships. For example, in conversation sequences, conversations between different people are interdependent; In music sequences, different rhythmic parts interact with each other; In video sequences, every frame of a video is a picture, and different objects in the picture have their own causal relationships. From another perspective, causality is more continuous in video sequences than in conversation sequences, the nature of dialogue leads to discrete causality.  

Through the arrangement and comparison of MVD, BVD and IVD, we found that the similarities between them lie in they all involve the same causal issues and conclusions. For example, there are priors in these data sets; All these datasets may have confounders, the confounders in them all exist in fork-structure, and all need to exclude the influence of confounders; Likewise, we can all carry out some causal operations such as interventions and counterfactual on these data sets. On the other hand, there are also some differences between them. In MVD, multiple variables are definite reflected in the number, meaning and sampling, therefore, when we can directly obtain the unique causal skeleton according to the priori of MVD and further construct the unique DSCM; In BVD, it has only two definite original variables, multidimensional time-space series data is labeled into multiple types, the causal skeleton can only be constructed by dividing other exogenous variables from the data according to the priori of researchers. Although there are a definite number of causal variables, the SSCM we finally get is not unique due to the weak prior of BVD. Besides, IVD also has two variables, but its data and labels exist in the form of sequence, so it is difficult for us to obtain other definite information except for the sample size. We need to take different sequence fragments as causal variables to construct the causal skeleton, so the USCM obtained is not uniquely caused by its non-samplability, and the number of USCM is close to the sample size of IVD. For MVD and BVD, there are plenty of ways to build SCMs from data sets, but there are gaps in IVD, and that's what we'll focus on in the next section. 

\section{Future Roadmaps}\label{section5}
This section focuses on some roadmaps of corresponding solutions for causal discovery in infinite-variable paradigms. Because the nodes of each sample in IVD are different, the sparsity of its sample space leads to the non-samplability of USCM. Therefore, the methods used in the construction of DSCM and SSCM cannot be continued here. Transforming non-samplability to samplability is the key to extending the causal discovery algorithm to undefinite task. In section \emph{A}, we proposed a priori-based causal discovery method, it is stipulated by the prior knowledge of each node is the same, so that the partial cause and effect diagram constituted from each node and its related nodes can be sampled; In section \emph{B}, we proposed sampling-based method, aiming to make every sample can generate enough and similar sample, to make the samples can be sampled; In section \emph{C}, we also proposed deterministic-based method, through mapping all nodes to a maximum graph, then the maximum graph can meet the samplability. We'll cover these roadmaps in more detail below.

\subsection{Prior-based Method}

\begin{figure}
  \includegraphics[width=1\linewidth]{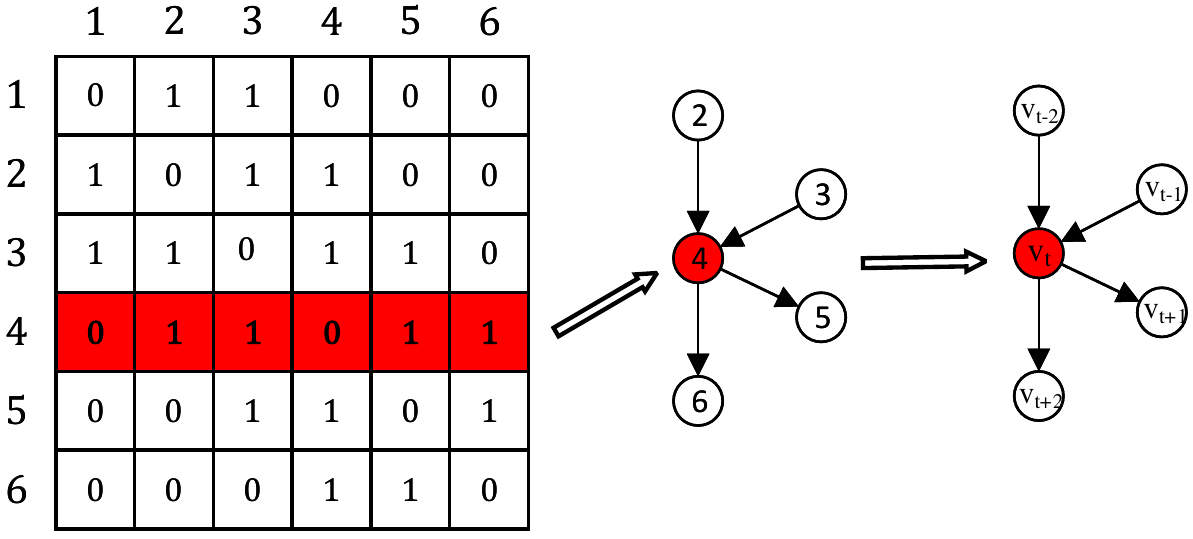}
  \caption{An explanation of how prior-based method constructs the 
  whole skeleton. The target node is inked by red.}
  \label{prior}
\end{figure}

In Section~\ref{undef}, we proposed support of non-sampling in IVD, which 
leads to an intractable injective model from variables to nodes. Specifically, 
as their datapoints are variable-length and unstructured, IVD segments 
differ from the MVD variables. Hence, instead of focusing on a common 
skeleton for all variable-length IVD sequences, we design a way to 
build a common skeleton for each segment from several broadly accepted 
prior knowledge, called prior-based method. In Figure~\ref{prior}, we 
detailed how to build a whole skeleton via prior-based sub-skeleton. Each 
sub-skeleton centered a target node and contained other nodes directly 
related to the target node. That is, each segment has a corresponding 
sub-skeleton via regarding the segment as the target node. Therefore, 
we can obtain a whole skeleton consisting of $N$ sub-skeleton, and $N$ 
equals the number of segment in a sequence. 

We can formalize the prior-based method under several mild hypotheses. 
We detailed 3 basic hypotheses as follows: 

\textbf{Hypothesis 1.} Each IVD segment has the same sub-skeleton. 

\textbf{Hypothesis 2.} As the target node, each segment is influenced 
by the fixed number of related segments. 

\textbf{Hypothesis 3.} There is no confounder in the final skeleton. 

Under \textbf{Hypothesis 1}, there is a common sub-skeleton for all 
segments in IVDs, and under \textbf{Hypothesis 2}, each target segment 
enjoys the complete independent relationships in corresponding sub-skeleton. 
Therefore, we can analysis the causal relationship in sub-skeleton with 
it satisfying the sampling. Besides, \textbf{Hypothesis 3} impose strict 
constraints upon designing sub-skeletons which ensure the whole skeleton 
not involved with confounder. 

Recent years, there are some conversation-based studies proposed several 
instantiation of sub-skeleton to achieve dialogue dataset problems. We 
show 3 generally accpeted prior knowledge which can be extended to other 
types of IVDs as follows: 

\textbf{Prior Knowledge 1.} \textit{Each target node recieves information from 
all other nodes. } 

Under \textbf{Prior Knowledge 1}, we can obtain a full-connected DAG as 
a sub-skeleton, all edges pointing to the target nodes. Moreover, the 
whole skeleton is a full-connected undirected graph. This prior offer 
that all variables are potentially related and need more complex causal 
discovery algorithm. \cite{wei-etal-2020-effective} models the clause-to-clause 
relationships via exploiting graph attention to learn clause representation, 
facilitating pair ranking through building the underlying relationship 
between two clauses. Besides, they proposed support of the causal relationship 
of inter-clauses that the probability of two distant clauses containing 
causal relationship is relatively small. Based on this consideration, 
they build a full-connected graph with each node representing a clause 
and each edge representing the latent relationship between two clauses. 
However, when the graph attention algorithm was used for yielding clause 
representation, they considered the relative position information between 
two clauses of a pair to cut off most edges. Furthermore, \cite{chen-etal-2020-end} 
proposed another full-connected skeleton for the same task. They construct a 
pair graph to model dependency relationships among local neighborhood candidate pairs. 
There are three categories of dependecy relationships and each dependency 
relationship representation a different way of information passing. 

\textbf{Prior Knowledge 2.} \textit{Each target node recieves information by 
different types of edges.}

Under \textbf{Prior Knowledge 2}, we can obtain a heterogeneous graph as 
a sub-skeleton, different types of edges pointing to the target nodes. Moreover, the 
whole skeleton is a heterogeneous undirected graph. By leveraging different 
types of edges, some strong inductive bias involved with the sub-skeletion. 
\cite{shen-etal-2021-directed} proposed that speaker identity is the 
strong inductive bias in conversation text. Hence they defined two types 
of edges: one is that two utterances belong to the same speaker, other is 
that two utterances has different speaker identities. Specifically, 
they proposed a more intuitive way to model the information flow between 
same and different speakers. In music sequence dataset, there are also 
similar studies which disentangle music sequence into different types. Such as 
pitch and rhythm \cite{luo2020mg}. Moreover, \cite{li2019regional} proposed 
support that regions are also an effective inductive bias when analyse 
music sequence.

\textbf{Prior Knowledge 3.} \textit{Each target node only recieves information from 
predecessor nodes.}

Under \textbf{Prior Knowledge 3}, we can obtain a acyclic graph as 
a sub-skeleton, all edges pointing to the successor nodes. Moreover, the 
whole skeleton is a PDAG graph. Under this strict prior knowledge, each 
node can not propagate information cackward to itself and its predecessors 
through any path. This characteristic indicates that a subskeleton can be regarded 
as a DAG. \cite{lian2021decn} provided supports of the emotional inertia, which 
was a concept that the interlocutor's emotional state was resistant to change, 
despite the influence from counterparts. Hence, they proposed a connectionist 
model that capture the dependence of graph via message passing from the 
predecessor nodes. Besides, \cite{chen2022learning} formalize the conversation 
via GNN model and describe the clause-to-clause relationship by DAG. 

Despite many studies and prior knowledges, there more problems applying 
them to USCM in IVDs. Confounder is the most crucial concerning when 
more than one prior knowledges above are adopted. To alleviate the bias 
lead by confounder, some excessive edges should be removed. Hence, how to 
define a more mild and effective prior to ensure the low confounder deserves 
more future research. 

\subsection{Sampling-based Method}
In order to solve the problem that the USCM is difficult to figure out, one solution is to increase the number of samples. If each node can perform enough samples, the relationship between nodes will be more robust. In order to augment the data, we can use some data augmentation method to sample.

\textbf{Condition 1.} The specific category to which each piece of data belongs can be known.

This condition is very strong, and we can regard the original data as incomplete data, so the idea of completing missing data can be adopted to obtain the relationship of complete data from original data\cite{adel2017learning}. Specifically, the missing mode of the data can be assumed according to the missing at random (MAR), and the filling mode can be the EM algorithm\cite{wei1990monte} or Gibbs sampling\cite{geman1984stochastic}.

\textbf{Condition 2.} The subject category of some data, and the upper limit of the category can be known.

In this condition, we can get an upper limit on the number of nodes in the causal graph\cite{tu2019causal}, so the method of augmenting the same theme can be adopted, and each node can be augmented separately, and then iteratively judge the mutual causal relationship according to the method of generating the causal graph, so as to complete the data augmentation.

\textbf{Condition 3.} The type of each piece of data in the data set can be specified, but each piece of data belongs to different nodes.

As long as the type of data can be clarified, it can still be augmented by data augmentation. Since there is only a single piece of data per node, data augmentation methods may rely on deeper, more incomprehensible connections to the data. A feasible idea is to use neural network models. There are some examples to prove it.

(1) Using the autoencoder (AE). The encoder uses a multi-layer with a small parameter network to learn, trying to learn its deep features, and uses the decoder to generate new data similar to the original data. Using the network to iterate, you can repeatedly augment the data, and then use other algorithms for causal discovery.

(2) Using the variational autoencoder (VAE). Due to its relatively strong generalization, it may be necessary to use a network with smaller parameters for learning. The decoder of the trained model can generate multiple sets of approximate data at one time, eliminating the error accumulation of the iterative process., to ensure that the distance between the newly generated data and the original data is close.

(3) Using the generative adversarial model GAN\cite{wang2020causal}. the core is to design a generator and an evaluator. In the generator, it is specified that it is generated according to a causal graph model, and in the evaluator, the generated data is evaluated according to the relationship with the original data by whether the original distribution has been changed. if the generated data and the original data are within a reasonable distance, the relationship can be regarded as consistent, otherwise inconsistent.

\textbf{Condition 4.} The category to which each piece of data in the dataset belongs is completely unknown.

This type of problem is probably the most difficult problem to solve, which means that the number of types of data and the relationship between data of the same type cannot even be known at all, so even these contents need to be self-generated from the data and optimized. One idea is to use the diffusion model DM to form a series of Markov chains of noised data by adding noise to the original data set, and then use its reverse process to generate data from isotropic Gaussian noise. There is no guarantee that the distribution of the new data set is exactly the same as that of the original data set, but the nature of the original data set itself is not enough to extract more information without a priori hypothesis, so it even seems impossible to judge whether there is a judgment. But more data is usually useful for subsequent learning, and maybe this way can help us find some missing data set construction.

\subsection{Maximum Graph Based Method}
In addition to the above two roadmaps, we can also try to propose a maximum complete graph based on the dataset or batch. From the properties of the Undefinite Structural Causal Model (USCM), we can know that for different samples in an IVD, the causal skeleton (causal diagram) we finally obtain is different. Therefore, we expect to construct a common graph of all samples, so that each sample can be mapped to this maximum graph. In this way, for each variable (sequence fragment) of each sample in the IVD, we can find a unique corresponding node in the maximum graph to represent, and each sample also has the same node. This is in line with the characteristics of our previously proposed definite structural causal model (DSCM), and the maximum graph based method actually turns infinite variables into definite variables. Therefore, after obtaining the causal skeleton according to this method, we can continue to use the causal discovery method used in the definite task to explore the causal relationship between variables.

\begin{figure}
  \includegraphics[width=0.95\linewidth]{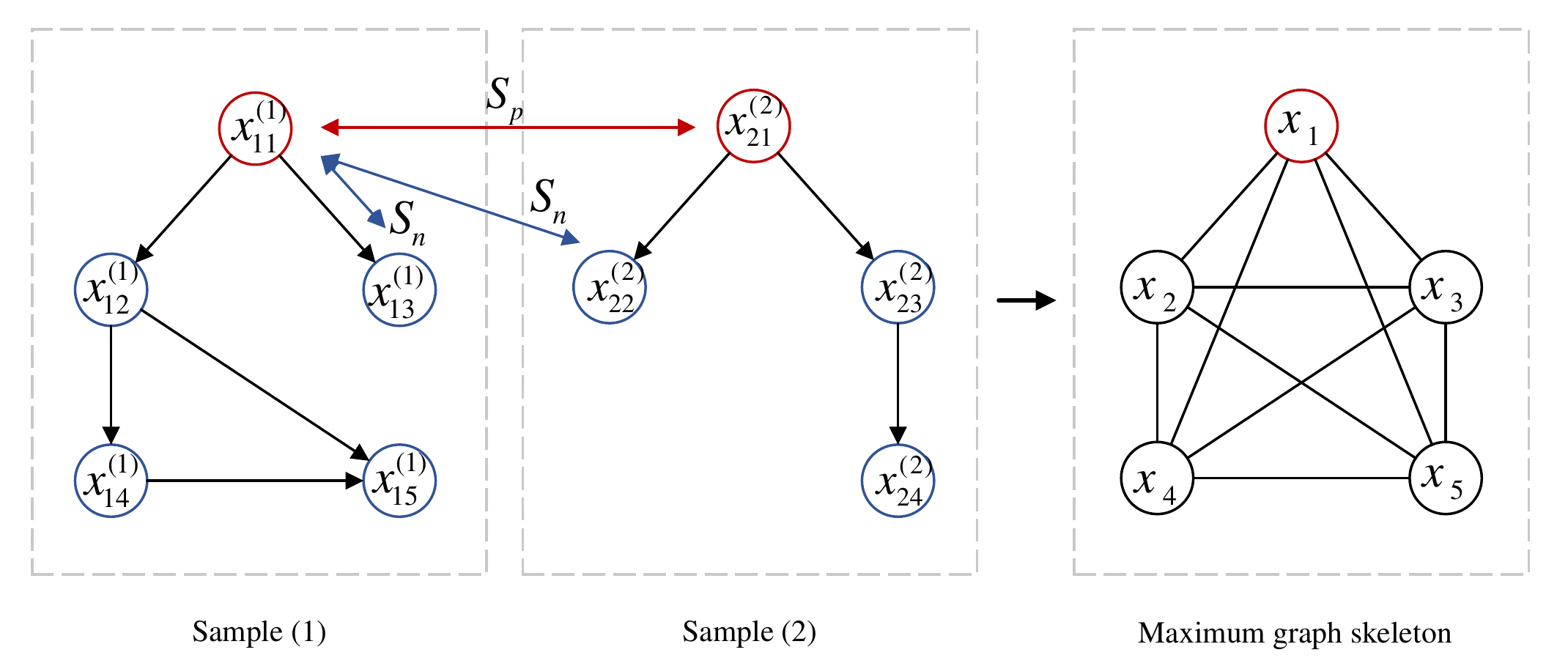}
  \caption{An Example of constructing maximum graph with two different samples in IVD. $x_{11}^{(1)}$ and $x_{21}^{(1)}$ are are mutually positive, their mapping in maximum gragh is $x_1$, the nodes in blue are their negatives. }
  \label{maximum}
\end{figure}

For any IVD, let it contain samples $(x_{11}^{(1)},x_{12}^{(1)},…,x_{15}^{(1)})$ and $(x_{21}^{(2)},x_{22}^{(2)},…,x_{24}^{(2)})$, the causal skeleton is shown in the Figure~\ref{maximum}. For any anchor in each sample skeleton, such as $x_{11}^{(j)}$, we can find the most similar nodes from other sample skeletons $x_{11}^{(1)},x_{21}^{(2)},… ,x_{(j-1)1}^{(j-1)}$, also known as a positive of $x_{11}^{(j)}$ , the distance between these similar nodes is very small, and eventually all can be represented as node $x_1$ in the maximum graph. Each anchor $x_{(j_1)}^{(j)}$ are not similar with other variables $x_{j_2}^{(j)}$ ,$x_{j_3}^{(j)}$,…,$x_{j_l}^{(j)} $in the same sample and variables in different samples except positive, the distance between them is large enough, and finally represented as different nodes $x_1,x_2,…,x_l$ in the common maximum graph.

The premise of constructing the maximum graph is to conduct sample mining\cite{kaya2019deep}, distinguishing positive and negative examples to improve the ability of models to recognize different variables and prevent overfitting from falling into local optimum, resulting in wrong skeletons. Therefore, we assume in an IVD, a typical set of samples consists of an anchor, negative $A$, negative $B$, and positive. Positive is the variables similar to the anchor variable, and negative A and B are variables different from the anchor. For the determination of positive and negative, we will make the following assumptions to ensure the representativeness of the same node for similar variables and to ensure that dissimilar variables can be effectively distinguished.

\textbf{Hypothesis 4.} The anchor in each sample can find a unique similar variable in other samples, which are recorded as positive.

\textbf{Hypothesis 5.} The anchor in each sample is dissimilar to other variables except itself, and the negative in these same samples$\{x_{j_1}^{(j)}, x_{j_2}^{(j)},…,x_{j_l}^{(j)}\}\setminus x_{anchor}^{(j)} $ are recorded as negative $A$.

\textbf{Hypothesis 6.} The anchor in each sample is dissimilar to the non-positive ones in other samples. These negatives in other samples $\{X_{j_1}, X_{j_2},…,X_{j_l}\}\setminus \{ X_{anchor}, X_{positive} \} $ are denoted as negative $B$.

As shown in Figure~\ref{maximum}, in light of the above three Hypothesises, we need to perform sample mining according to the distance between the anchor and its positive and negative pairs. This distance can be calculated by the Euclidean distance\cite{danielsson1980euclidean} or Mahalanobis distance\cite{de2000mahalanobis} of direct metric learning\cite{kulis2013metric}. The distance between anchor and positive is denoted as $S_p$, and the distance between anchor and negative $A$ and $B$ is denoted as $S_n$. According to the difference of distance between the anchor and its positive and negative, we can identify the variables with the closest distance to the anchor as its positive according to Hard Negative Mining\cite{bucher2016hard}, and the rest of the samples are its negative, so there is $S_p \textless S_n$. For the maximum number of nodes in the graph, we can prior set it to be equal to the maximum number of variables in a single sample, expressed as $l = max\{l^{(1)}, l^{(2)},…, l^{(k)}\} $, where k is the sample size of IVD, and $l^{(k)}$ is the number of variables in the $k$ th sample.

Loss functions also play a very important role in deep metric learning. Many loss functions for deep metric learning are built on sample pairs or triplets, and the magnitude of the sample space is very large. In order to avoid falling into a local optimum, we usually do targeted optimization on the sample. The input of Contrastive loss\cite{wang2021understanding} is a sample pair consisting of two samples, and the input of Triplet loss\cite{ge2018deep} consists of a triplet containing the anchor with its positive and negative. The input in the maximum graph we constructed involves anchors, positives, and multiple negatives from the same sample and different samples. There are many samples to be compared, and it is not suitable to use Contrastive loss and Triplet loss directly. Therefore, we summarize here two loss functions suitable for such multiple inputs:

(1)\textbf{N-pair loss.}\cite{wang2017deep, sohn2016improved} It uses multiple negative samples when calculating the loss function value of mini-batch, and does not slow down like ternary loss and contrastive loss. Each training sample of N-pair loss consists of $N + 2$ tuples, namely $x, x^+, x_1,...,x_k$, where $x^+$ represents the positive relative to anchor $x, x_1,...,x_k$ are negatives respectively picked from current sample and other samples.

\begin{center}
\begin{equation}
L(x, x^+, x_1,...,x_k) = log(1+\sum_{i = 1}^{K}exp(f^Tf_i - f^Tf^+))
\end{equation}
\end{center}

$f(\cdot;\theta )$ is the embedding kernel defined by the deep neural network.

(2)\textbf{Lifted Structured loss.}\cite{oh2016deep,he2019deep} The loss is calculated based on all positive and negative sample pairs in the training set (mini batch), which is defined

\begin{center}
\begin{equation}
L = \frac{1}{2| \hat{P}|  } \sum_{{(i,j)}\in \hat{P}}\max{(0,L_{i,j})}^2
\end{equation}
\end{center}

\begin{center}
\begin{equation}
L_{i,j} = \max(\max\limits_{(i,k)\in \hat{N}}\alpha - D_{i,k}, \max\limits_{(j,l)\in \hat{N} }\alpha - D_{j,l}) + D_{i,j}
\end{equation}
\end{center}

Among them, $\hat{P}$ is the set of positive sample pairs; $\hat{N}$ is the set of negative sample pairs;$ D_{i,j}$ is the distance of the sample pair $(i,j)$.

During this learning process, if without informative sample mining or an select an inappropriate loss function, the problem of sample inseparable\cite{liu2010study} may occur. The inseparable problem means that no matter what value $\theta $ of the projection function takes, it cannot satisfy $d(f_\theta  (x_i),f_\theta  (x_j ))  \textless  d(f_\theta  (x_i),f_\theta  (x_l))$, where $x_j$ is a positive of $x_i$, $x_l$ is the negative of $x_i$. This situation exists in both linear metric learning and deep metric learning, because most loss functions focus on minimizing $(S_n-S_p)$, it is inflexible because it penalizes every single similarity score equally. At this time, we need to use a more effective metric learning algorithm, which can eliminate indivisible samples without affecting the stability of the loss function. Circle loss\cite{sun2020circle} proposes if a similarity score is far from the optimal center, it should be paid more attention (i.e., penalized). For this purpose, circle loss re-weights those under-optimized similarity scores.

\begin{center}
\begin{equation}
L_{circle} = \log [1+\sum_{i = 1}^{K} \sum_{j = 1}^{L}\exp (\gamma \alpha _n^j(s_n^j - \Delta_n) - \alpha _p^j(s_p^j - \Delta_p))]
\end{equation}
\end{center}
\label{circle}

In Equation~\ref{circle}, $s_n^j = d(x,x_n^j)$ is the distance between the anchor x and its $j$ th negative, $s_p^j = d(x,x_p^j)$ is the distance between the anchor x and its $j$ th positive, $\alpha _n^j$ = ${[s_n^j-O_n]}_+$ and $\alpha _p^j$ = ${[O_p-s_p^j]}_+$ are the parameters obtained by self-learning. The loss function of Cirlce Loss can be transformed into 

\begin{center}
\begin{equation}
L_{circle} = \log [1+\sum_{i = 1}^{K} \sum_{j = 1}^{L}\exp (\gamma (\hat{s}_n^j- \hat{s}_p^j + m))]
\end{equation}
\end{center}
\label{circle1}

Where $m = \alpha _p^j\Delta_p - \alpha _n^j \Delta_n$ can be considered as a margin, $\hat{s}_n^j = \alpha _n^js_n^j$ and $\hat{s}_p^j = \alpha _p^js_p^j$ are weighted distances, we minimize $(\hat{s}_n^j- \hat{s}_p^j + m)$ to satisfy constraint $\hat{s}_n^j \textgreater \hat{s}_p^j$. Compared with the traditional optimization $s$ and the weighted $\hat{s} $, there still has the same gradient and decision surface at different sample points, but it can dynamically adjust the gradient to make the optimization direction clearer.

The above is one of roadmaps for how to build USCM from the perspective of constructing a maximum graph. After obtaining the skeleton of the maximum graph, the researcher also needs to analyze the different included sample variables, so as to name the nodes in the maximum graph. How to effectively learn the loss function, reduce the computational complexity, and how to determine the maximum number of nodes in the graph are issues that need to be further considered in practice. Simply equating the maximum number of nodes with the maximum number of nodes in a single sample may result in some samples not being able to find a suitable mapping in the ending maximum graph. Therefore, we may need to combine the loss function and set a threshold to limit the number of nodes in the maximum graph, so as to ensure that each variable in a single sample can find a reasonable mapping in the maximum graph.

\section{Conclusion}
Our work takes different variable paradigms as the entry point, defining and dividing the possible tasks of causal discovery into three types, namely definite task, semidefinite task, and undefinite task. Under each task, we give a specific definition of the task, and correspondingly define the dataset applied to the task, and the final manifestation of them: the structural causal models(SCMs) of different paradigms. In the definite task, we define the multi-variable dataset (MVD) and summarize some common MVDs, and then define the definite structural causal model (DSCM) according to the characteristics of the task and dataset, summarizing the existing applicable methods under this task in various. In the semidefinite task, we define the binary-variable dataset (BVD) and the semidefinite structural causal model (SSCM), summarize the differences of semidefinite tasks and semidefinite structural causal models between definite variable paradigm and their characteristics, and also some BVDs in different domains and its causal discovery methods. In the undefinite task, we give the definition of the infinite-variable dataset (IVD) and the undefinite structural causal model (USCM), clarifying the characteristics of this variable paradigm and its differences of tasks and SCMs between the first two variable paradigms. Undefinite tasks are the existing research gap in the field of causal discovery, and it is also a part that will inevitably be involved in causal discovery with the rapid development of deep learning, because causality itself is a more advanced and perfect sequence relationship. Here we propose three roadmaps from different perspectives, namely prior-based method, sampling-based method and maximum graph based method, in order to solve the problem of lack of sufficient sampling in undefinite tasks, which makes it difficult to build USCM, and how to effectively implement it also depends on further exploration attempts by researchers. In addition, there are many special challenges in the process of reliable causal discovery, such as sampling bias, measurement error, the presence of confounders, nonlinear effects, etc., these are also problems that we need to solve in improving the quality of causal structure search.

\bibliographystyle{IEEEtran}
\bibliography{custom}

\begin{thebibliography}{100}
\providecommand{\url}[1]{#1}
\csname url@samestyle\endcsname
\providecommand{\newblock}{\relax}
\providecommand{\bibinfo}[2]{#2}
\providecommand{\BIBentrySTDinterwordspacing}{\spaceskip=0pt\relax}
\providecommand{\BIBentryALTinterwordstretchfactor}{4}
\providecommand{\BIBentryALTinterwordspacing}{\spaceskip=\fontdimen2\font plus
\BIBentryALTinterwordstretchfactor\fontdimen3\font minus
  \fontdimen4\font\relax}
\providecommand{\BIBforeignlanguage}[2]{{%
\expandafter\ifx\csname l@#1\endcsname\relax
\typeout{** WARNING: IEEEtran.bst: No hyphenation pattern has been}%
\typeout{** loaded for the language `#1'. Using the pattern for}%
\typeout{** the default language instead.}%
\else
\language=\csname l@#1\endcsname
\fi
#2}}
\providecommand{\BIBdecl}{\relax}
\BIBdecl

\bibitem{pearl2009causality}
J.~Pearl, \emph{Causality}.\hskip 1em plus 0.5em minus 0.4em\relax Cambridge
  university press, 2009.

\bibitem{cox1992causality}
D.~R. Cox, ``Causality: some statistical aspects,'' \emph{Journal of the Royal
  Statistical Society: Series A (Statistics in Society)}, vol. 155, no.~2, pp.
  291--301, 1992.

\bibitem{heckman2022causality}
J.~J. Heckman and R.~Pinto, ``Causality and econometrics,'' National Bureau of
  Economic Research, Tech. Rep., 2022.

\bibitem{eells1991probabilistic}
E.~Eells, \emph{Probabilistic causality}.\hskip 1em plus 0.5em minus
  0.4em\relax Cambridge University Press, 1991, vol.~1.

\bibitem{berzuini2012causality}
C.~Berzuini, P.~Dawid, and L.~Bernardinell, \emph{Causality: Statistical
  perspectives and applications}.\hskip 1em plus 0.5em minus 0.4em\relax John
  Wiley \& Sons, 2012.

\bibitem{hicks1980causality}
J.~Hicks \emph{et~al.}, \emph{Causality in economics}.\hskip 1em plus 0.5em
  minus 0.4em\relax Australian National University Press, 1980.

\bibitem{heckman2008econometric}
J.~J. Heckman, ``Econometric causality,'' \emph{International statistical
  review}, vol.~76, no.~1, pp. 1--27, 2008.

\bibitem{scholkopf2022causality}
B.~Sch{\"o}lkopf, ``Causality for machine learning,'' in \emph{Probabilistic
  and Causal Inference: The Works of Judea Pearl}, 2022, pp. 765--804.

\bibitem{xu2020causality}
G.~Xu, T.~D. Duong, Q.~Li, S.~Liu, and X.~Wang, ``Causality learning: a new
  perspective for interpretable machine learning,'' \emph{arXiv preprint
  arXiv:2006.16789}, 2020.

\bibitem{meliou2010causality}
A.~Meliou, W.~Gatterbauer, J.~Y. Halpern, C.~Koch, K.~F. Moore, and D.~Suciu,
  ``Causality in databases,'' \emph{IEEE Data Engineering Bulletin}, vol.~33,
  no. ARTICLE, pp. 59--67, 2010.

\bibitem{raynal1996logical}
M.~Raynal and M.~Singhal, ``Logical time: Capturing causality in distributed
  systems,'' \emph{Computer}, vol.~29, no.~2, pp. 49--56, 1996.

\bibitem{vlontzos2022review}
A.~Vlontzos, D.~Rueckert, and B.~Kainz, ``A review of causality for learning
  algorithms in medical image analysis,'' \emph{arXiv preprint
  arXiv:2206.05498}, 2022.

\bibitem{gottlieb2002relational}
G.~Gottlieb and C.~T. Halpern, ``A relational view of causality in normal and
  abnormal development,'' \emph{Development and psychopathology}, vol.~14,
  no.~3, pp. 421--435, 2002.

\bibitem{castro2020causality}
D.~C. Castro, I.~Walker, and B.~Glocker, ``Causality matters in medical
  imaging,'' \emph{Nature Communications}, vol.~11, no.~1, pp. 1--10, 2020.

\bibitem{child1994causality}
W.~Child, \emph{Causality, interpretation, and the mind}.\hskip 1em plus 0.5em
  minus 0.4em\relax Clarendon Press, 1994.

\bibitem{young2022development}
G.~Young, ``Development, stages, and causality,'' in \emph{Causality and
  Neo-Stages in Development}.\hskip 1em plus 0.5em minus 0.4em\relax Springer,
  2022, pp. 21--42.

\bibitem{schield1995correlation}
M.~Schield, ``Correlation, determination and causality in introductory
  statistics,'' \emph{American Statistical Association, Section on Statistical
  Education}, 1995.

\bibitem{simon2017spurious}
H.~A. Simon, ``Spurious correlation: A causal interpretation,'' in \emph{Causal
  models in the social sciences}.\hskip 1em plus 0.5em minus 0.4em\relax
  Routledge, 2017, pp. 7--22.

\bibitem{boutaba2018comprehensive}
R.~Boutaba, M.~A. Salahuddin, N.~Limam, S.~Ayoubi, N.~Shahriar,
  F.~Estrada-Solano, and O.~M. Caicedo, ``A comprehensive survey on machine
  learning for networking: evolution, applications and research
  opportunities,'' \emph{Journal of Internet Services and Applications},
  vol.~9, no.~1, pp. 1--99, 2018.

\bibitem{smith2022real}
J.~J. Smith, S.~Amershi, S.~Barocas, H.~Wallach, and J.~Wortman~Vaughan, ``Real
  ml: Recognizing, exploring, and articulating limitations of machine learning
  research,'' in \emph{2022 ACM Conference on Fairness, Accountability, and
  Transparency}, 2022, pp. 587--597.

\bibitem{guo2020survey}
R.~Guo, L.~Cheng, J.~Li, P.~R. Hahn, and H.~Liu, ``A survey of learning
  causality with data: Problems and methods,'' \emph{ACM Computing Surveys
  (CSUR)}, vol.~53, no.~4, pp. 1--37, 2020.

\bibitem{series2010guide}
I.~T. Series and D.~A. Graph, ``A guide on data analysis,'' 2010.

\bibitem{sauer2013use}
B.~Sauer and T.~J. VanderWeele, ``Use of directed acyclic graphs,'' in
  \emph{Developing a protocol for observational comparative effectiveness
  research: a user's guide}.\hskip 1em plus 0.5em minus 0.4em\relax Agency for
  Healthcare Research and Quality (US), 2013.

\bibitem{vowels2021d}
M.~J. Vowels, N.~C. Camgoz, and R.~Bowden, ``D'ya like dags? a survey on
  structure learning and causal discovery,'' \emph{ACM Computing Surveys
  (CSUR)}, 2021.

\bibitem{gelman2014understanding}
A.~Gelman, J.~Hwang, and A.~Vehtari, ``Understanding predictive information
  criteria for bayesian models,'' \emph{Statistics and computing}, vol.~24,
  no.~6, pp. 997--1016, 2014.

\bibitem{spirtes2016causal}
P.~Spirtes and K.~Zhang, ``Causal discovery and inference: concepts and recent
  methodological advances,'' 2016.

\bibitem{2018The}
J.~Pearl and D.~Mackenzie, ``The book of why : the new science of cause and
  effect,'' \emph{Science}, vol. 361, no. 6405, pp. 855.2--855, 2018.

\bibitem{eberhardt2017introduction}
F.~Eberhardt, ``Introduction to the foundations of causal discovery,''
  \emph{International Journal of Data Science and Analytics}, vol.~3, no.~2,
  pp. 81--91, 2017.

\bibitem{cooper2013causal}
G.~F. Cooper and C.~Yoo, ``Causal discovery from a mixture of experimental and
  observational data,'' \emph{arXiv preprint arXiv:1301.6686}, 2013.

\bibitem{cinelli2019sensitivity}
C.~Cinelli, D.~Kumor, B.~Chen, J.~Pearl, and E.~Bareinboim, ``Sensitivity
  analysis of linear structural causal models,'' in \emph{International
  conference on machine learning}.\hskip 1em plus 0.5em minus 0.4em\relax PMLR,
  2019, pp. 1252--1261.

\bibitem{bongers2021foundations}
S.~Bongers, P.~Forr{\'e}, J.~Peters, and J.~M. Mooij, ``Foundations of
  structural causal models with cycles and latent variables,'' \emph{The Annals
  of Statistics}, vol.~49, no.~5, pp. 2885--2915, 2021.

\bibitem{louizos2017causal}
C.~Louizos, U.~Shalit, J.~M. Mooij, D.~Sontag, R.~Zemel, and M.~Welling,
  ``Causal effect inference with deep latent-variable models,'' \emph{Advances
  in neural information processing systems}, vol.~30, 2017.

\bibitem{pearl2010causal}
J.~Pearl, ``Causal inference,'' \emph{Causality: objectives and assessment},
  pp. 39--58, 2010.

\bibitem{pearl2009causal}
------, ``Causal inference in statistics: An overview,'' \emph{Statistics
  surveys}, vol.~3, pp. 96--146, 2009.

\bibitem{ding2020reliable}
R.~Ding, Y.~Liu, J.~Tian, Z.~Fu, S.~Han, and D.~Zhang, ``Reliable and efficient
  anytime skeleton learning,'' in \emph{Proceedings of the AAAI Conference on
  Artificial Intelligence}, vol.~34, no.~06, 2020, pp. 10\,101--10\,109.

\bibitem{stone2004independent}
J.~V. Stone, ``Independent component analysis: a tutorial introduction,'' 2004.

\bibitem{0Graphical}
T.~S. Verma, ``Graphical aspects of causal models.''

\bibitem{spirtes2000causation}
P.~Spirtes, C.~N. Glymour, R.~Scheines, and D.~Heckerman, \emph{Causation,
  prediction, and search}.\hskip 1em plus 0.5em minus 0.4em\relax MIT press,
  2000.

\bibitem{rosa2011inferring}
G.~J. Rosa, B.~D. Valente, G.~de~los Campos, X.-L. Wu, D.~Gianola, and M.~A.
  Silva, ``Inferring causal phenotype networks using structural equation
  models,'' \emph{Genetics Selection Evolution}, vol.~43, no.~1, pp. 1--13,
  2011.

\bibitem{chicharro2014algorithms}
D.~Chicharro and S.~Panzeri, ``Algorithms of causal inference for the analysis
  of effective connectivity among brain regions,'' \emph{Frontiers in
  neuroinformatics}, vol.~8, p.~64, 2014.

\bibitem{ramsey2012adjacency}
J.~Ramsey, J.~Zhang, and P.~L. Spirtes, ``Adjacency-faithfulness and
  conservative causal inference,'' \emph{arXiv preprint arXiv:1206.6843}, 2012.

\bibitem{uhler2013geometry}
C.~Uhler, G.~Raskutti, P.~B{\"u}hlmann, and B.~Yu, ``Geometry of the
  faithfulness assumption in causal inference,'' \emph{The Annals of
  Statistics}, pp. 436--463, 2013.

\bibitem{kalisch2007estimating}
M.~Kalisch and P.~B{\"u}hlman, ``Estimating high-dimensional directed acyclic
  graphs with the pc-algorithm.'' \emph{Journal of Machine Learning Research},
  vol.~8, no.~3, 2007.

\bibitem{chickering2002optimal}
D.~M. Chickering, ``Optimal structure identification with greedy search,''
  \emph{Journal of machine learning research}, vol.~3, no. Nov, pp. 507--554,
  2002.

\bibitem{shimizu2006linear}
S.~Shimizu, P.~O. Hoyer, A.~Hyv{\"a}rinen, A.~Kerminen, and M.~Jordan, ``A
  linear non-gaussian acyclic model for causal discovery.'' \emph{Journal of
  Machine Learning Research}, vol.~7, no.~10, 2006.

\bibitem{shimizu2011directlingam}
S.~Shimizu, T.~Inazumi, Y.~Sogawa, A.~Hyv{\"a}rinen, Y.~Kawahara, T.~Washio,
  P.~O. Hoyer, and K.~Bollen, ``Directlingam: A direct method for learning a
  linear non-gaussian structural equation model,'' \emph{The Journal of Machine
  Learning Research}, vol.~12, pp. 1225--1248, 2011.

\bibitem{hoyer2008nonlinear}
P.~Hoyer, D.~Janzing, J.~M. Mooij, J.~Peters, and B.~Sch{\"o}lkopf, ``Nonlinear
  causal discovery with additive noise models,'' \emph{Advances in neural
  information processing systems}, vol.~21, 2008.

\bibitem{peters2014causal}
J.~Peters, J.~M. Mooij, D.~Janzing, and B.~Sch{\"o}lkopf, ``Causal discovery
  with continuous additive noise models,'' 2014.

\bibitem{tsamardinos2006max}
I.~Tsamardinos, L.~E. Brown, and C.~F. Aliferis, ``The max-min hill-climbing
  bayesian network structure learning algorithm,'' \emph{Machine learning},
  vol.~65, no.~1, pp. 31--78, 2006.

\bibitem{cai2018self}
R.~Cai, J.~Qiao, Z.~Zhang, and Z.~Hao, ``Self: structural equational likelihood
  framework for causal discovery,'' in \emph{Proceedings of the AAAI Conference
  on Artificial Intelligence}, vol.~32, no.~1, 2018.

\bibitem{wong2002hybrid}
M.~L. Wong, S.~Y. Lee, and K.~S. Leung, ``A hybrid approach to discover
  bayesian networks from databases using evolutionary programming,'' in
  \emph{2002 IEEE International Conference on Data Mining, 2002.
  Proceedings.}\hskip 1em plus 0.5em minus 0.4em\relax IEEE, 2002, pp.
  498--505.

\bibitem{mittal2021affect2mm}
T.~Mittal, P.~Mathur, A.~Bera, and D.~Manocha, ``Affect2mm: Affective analysis
  of multimedia content using emotion causality,'' in \emph{Proceedings of the
  IEEE/CVF Conference on Computer Vision and Pattern Recognition}, 2021, pp.
  5661--5671.

\bibitem{zhu2020cookgan}
B.~Zhu and C.-W. Ngo, ``Cookgan: Causality based text-to-image synthesis,'' in
  \emph{Proceedings of the IEEE/CVF Conference on Computer Vision and Pattern
  Recognition}, 2020, pp. 5519--5527.

\bibitem{chen2021towards}
Y.~Chen, X.~Yang, T.-J. Cham, and J.~Cai, ``Towards unbiased visual emotion
  recognition via causal intervention,'' \emph{arXiv preprint
  arXiv:2107.12096}, 2021.

\bibitem{oh2021causal}
G.~Oh, E.~Jeong, and S.~Lim, ``Causal affect prediction model using a facial
  image sequence,'' \emph{arXiv preprint arXiv:2107.03886}, 2021.

\bibitem{rao2021counterfactual}
Y.~Rao, G.~Chen, J.~Lu, and J.~Zhou, ``Counterfactual attention learning for
  fine-grained visual categorization and re-identification,'' in
  \emph{Proceedings of the IEEE/CVF International Conference on Computer
  Vision}, 2021, pp. 1025--1034.

\bibitem{sridhar2019estimating}
D.~Sridhar and L.~Getoor, ``Estimating causal effects of tone in online
  debates,'' \emph{arXiv preprint arXiv:1906.04177}, 2019.

\bibitem{egami2018make}
N.~Egami, C.~J. Fong, J.~Grimmer, M.~E. Roberts, and B.~M. Stewart, ``How to
  make causal inferences using texts,'' \emph{arXiv preprint arXiv:1802.02163},
  2018.

\bibitem{zhang2020quantifying}
J.~Zhang, S.~Mullainathan, and C.~Danescu-Niculescu-Mizil, ``Quantifying the
  causal effects of conversational tendencies,'' \emph{Proceedings of the ACM
  on Human-Computer Interaction}, vol.~4, no. CSCW2, pp. 1--24, 2020.

\bibitem{imbens2020potential}
G.~W. Imbens, ``Potential outcome and directed acyclic graph approaches to
  causality: Relevance for empirical practice in economics,'' \emph{Journal of
  Economic Literature}, vol.~58, no.~4, pp. 1129--79, 2020.

\bibitem{pearl1988probabilistic}
J.~Pearl, \emph{Probabilistic reasoning in intelligent systems: networks of
  plausible inference}.\hskip 1em plus 0.5em minus 0.4em\relax Morgan kaufmann,
  1988.

\bibitem{glymour2019review}
C.~Glymour, K.~Zhang, and P.~Spirtes, ``Review of causal discovery methods
  based on graphical models,'' \emph{Frontiers in genetics}, vol.~10, p. 524,
  2019.

\bibitem{vanderweele2007directed}
T.~J. VanderWeele and J.~M. Robins, ``Directed acyclic graphs, sufficient
  causes, and the properties of conditioning on a common effect,''
  \emph{American journal of epidemiology}, vol. 166, no.~9, pp. 1096--1104,
  2007.

\bibitem{dietrich2019temporal}
H.~Dietrich, T.~Wolf, T.~Kawohl, J.~Wehberg, G.~K{\"a}ndler, T.~Mette,
  A.~R{\"o}der, and J.~B{\"o}hner, ``Temporal and spatial high-resolution
  climate data from 1961 to 2100 for the german national forest inventory
  (nfi),'' \emph{Annals of Forest Science}, vol.~76, no.~1, pp. 1--14, 2019.

\bibitem{van2006syntren}
T.~Van~den Bulcke, K.~Van~Leemput, B.~Naudts, P.~van Remortel, H.~Ma,
  A.~Verschoren, B.~De~Moor, and K.~Marchal, ``Syntren: a generator of
  synthetic gene expression data for design and analysis of structure learning
  algorithms,'' \emph{BMC bioinformatics}, vol.~7, no.~1, pp. 1--12, 2006.

\bibitem{web:US}
U.~D. of~Commerce, ``Website of the u.s. census bureau,'' 1994, uRL.
  \url{http://www.census.gov/.}

\bibitem{ristanoski2013discrimination}
G.~Ristanoski, W.~Liu, and J.~Bailey, ``Discrimination aware classification for
  imbalanced datasets,'' in \emph{Proceedings of the 22nd ACM international
  conference on Information \& Knowledge Management}, 2013, pp. 1529--1532.

\bibitem{iacoviello2019foreign}
M.~Iacoviello and G.~Navarro, ``Foreign effects of higher us interest rates,''
  \emph{Journal of International Money and Finance}, vol.~95, pp. 232--250,
  2019.

\bibitem{asuncion2007uci}
A.~Asuncion and D.~Newman, ``Uci machine learning repository,'' 2007.

\bibitem{huang2018generalized}
B.~Huang, K.~Zhang, Y.~Lin, B.~Sch{\"o}lkopf, and C.~Glymour, ``Generalized
  score functions for causal discovery,'' in \emph{Proceedings of the 24th ACM
  SIGKDD international conference on knowledge discovery \& data mining}, 2018,
  pp. 1551--1560.

\bibitem{solly2014factors}
E.~F. Solly, I.~Sch{\"o}ning, S.~Boch, E.~Kandeler, S.~Marhan, B.~Michalzik,
  J.~M{\"u}ller, J.~Zscheischler, S.~E. Trumbore, and M.~Schrumpf, ``Factors
  controlling decomposition rates of fine root litter in temperate forests and
  grasslands,'' \emph{Plant and Soil}, vol. 382, no.~1, pp. 203--218, 2014.

\bibitem{nash1994population}
W.~J. Nash, T.~L. Sellers, S.~R. Talbot, A.~J. Cawthorn, and W.~B. Ford, ``The
  population biology of abalone (haliotis species) in tasmania. i. blacklip
  abalone (h. rubra) from the north coast and islands of bass strait,''
  \emph{Sea Fisheries Division, Technical Report}, vol.~48, p. p411, 1994.

\bibitem{huang2020causal}
B.~Huang, K.~Zhang, J.~Zhang, J.~D. Ramsey, R.~Sanchez-Romero, C.~Glymour, and
  B.~Sch{\"o}lkopf, ``Causal discovery from heterogeneous/nonstationary data.''
  \emph{J. Mach. Learn. Res.}, vol.~21, no.~89, pp. 1--53, 2020.

\bibitem{sachs2005causal}
K.~Sachs, O.~Perez, D.~Pe'er, D.~A. Lauffenburger, and G.~P. Nolan, ``Causal
  protein-signaling networks derived from multiparameter single-cell data,''
  \emph{Science}, vol. 308, no. 5721, pp. 523--529, 2005.

\bibitem{tu2019causal}
R.~Tu, C.~Zhang, P.~Ackermann, K.~Mohan, H.~Kjellstr{\"o}m, and K.~Zhang,
  ``Causal discovery in the presence of missing data,'' in \emph{The 22nd
  International Conference on Artificial Intelligence and Statistics}.\hskip
  1em plus 0.5em minus 0.4em\relax PMLR, 2019, pp. 1762--1770.

\bibitem{wang2003training}
X.~Wang, R.~Hutchinson, and T.~M. Mitchell, ``Training fmri classifiers to
  detect cognitive states across multiple human subjects,'' \emph{Advances in
  neural information processing systems}, vol.~16, 2003.

\bibitem{baba2004partial}
K.~Baba, R.~Shibata, and M.~Sibuya, ``Partial correlation and conditional
  correlation as measures of conditional independence,'' \emph{Australian \&
  New Zealand Journal of Statistics}, vol.~46, no.~4, pp. 657--664, 2004.

\bibitem{prakasa2009conditional}
B.~L. Prakasa~Rao, ``Conditional independence, conditional mixing and
  conditional association,'' \emph{Annals of the Institute of Statistical
  Mathematics}, vol.~61, no.~2, pp. 441--460, 2009.

\bibitem{lauritzen2000causal}
S.~L. Lauritzen, ``Causal inference from,'' \emph{Complex stochastic systems},
  p.~63, 2000.

\bibitem{weinberger2018faithfulness}
N.~Weinberger, ``Faithfulness, coordination and causal coincidences,''
  \emph{Erkenntnis}, vol.~83, no.~2, pp. 113--133, 2018.

\bibitem{sadeghi2017faithfulness}
K.~Sadeghi, ``Faithfulness of probability distributions and graphs,''
  \emph{Journal of Machine Learning Research}, vol.~18, no. 148, pp. 1--29,
  2017.

\bibitem{andersson1997characterization}
S.~A. Andersson, D.~Madigan, and M.~D. Perlman, ``A characterization of markov
  equivalence classes for acyclic digraphs,'' \emph{The Annals of Statistics},
  vol.~25, no.~2, pp. 505--541, 1997.

\bibitem{cole2010illustrating}
S.~R. Cole, R.~W. Platt, E.~F. Schisterman, H.~Chu, D.~Westreich,
  D.~Richardson, and C.~Poole, ``Illustrating bias due to conditioning on a
  collider,'' \emph{International journal of epidemiology}, vol.~39, no.~2, pp.
  417--420, 2010.

\bibitem{kalisch2010pcalg}
M.~Kalisch, M.~M{\"a}chler, and D.~Colombo, ``pcalg: estimation of cpdag/pag
  and causal inference using the ida algorithm,'' \emph{URL http://CRAN.
  R-project. org/package= pcalg}, 2010.

\bibitem{pearl2000models}
J.~Pearl \emph{et~al.}, ``Models, reasoning and inference,'' \emph{Cambridge,
  UK: CambridgeUniversityPress}, vol.~19, no.~2, 2000.

\bibitem{colombo2012learning}
D.~Colombo, M.~H. Maathuis, M.~Kalisch, and T.~S. Richardson, ``Learning
  high-dimensional directed acyclic graphs with latent and selection
  variables,'' \emph{The Annals of Statistics}, pp. 294--321, 2012.

\bibitem{zhang2008completeness}
J.~Zhang, ``On the completeness of orientation rules for causal discovery in
  the presence of latent confounders and selection bias,'' \emph{Artificial
  Intelligence}, vol. 172, no. 16-17, pp. 1873--1896, 2008.

\bibitem{shen2020challenges}
X.~Shen, S.~Ma, P.~Vemuri, and G.~Simon, ``Challenges and opportunities with
  causal discovery algorithms: application to alzheimer's pathophysiology,''
  \emph{Scientific reports}, vol.~10, no.~1, pp. 1--12, 2020.

\bibitem{zhang2017causal}
K.~Zhang, B.~Huang, J.~Zhang, C.~Glymour, and B.~Sch{\"o}lkopf, ``Causal
  discovery from nonstationary/heterogeneous data: Skeleton estimation and
  orientation determination,'' in \emph{IJCAI: Proceedings of the Conference},
  vol. 2017.\hskip 1em plus 0.5em minus 0.4em\relax NIH Public Access, 2017, p.
  1347.

\bibitem{hauser2012characterization}
A.~Hauser and P.~B{\"u}hlmann, ``Characterization and greedy learning of
  interventional markov equivalence classes of directed acyclic graphs,''
  \emph{The Journal of Machine Learning Research}, vol.~13, no.~1, pp.
  2409--2464, 2012.

\bibitem{burnham2004multimodel}
K.~P. Burnham and D.~R. Anderson, ``Multimodel inference: understanding aic and
  bic in model selection,'' \emph{Sociological methods \& research}, vol.~33,
  no.~2, pp. 261--304, 2004.

\bibitem{neath2012bayesian}
A.~A. Neath and J.~E. Cavanaugh, ``The bayesian information criterion:
  background, derivation, and applications,'' \emph{Wiley Interdisciplinary
  Reviews: Computational Statistics}, vol.~4, no.~2, pp. 199--203, 2012.

\bibitem{suzuki2017theoretical}
J.~Suzuki, ``A theoretical analysis of the bdeu scores in bayesian network
  structure learning,'' \emph{Behaviormetrika}, vol.~44, no.~1, pp. 97--116,
  2017.

\bibitem{liu2012empirical}
Z.~Liu, B.~Malone, and C.~Yuan, ``Empirical evaluation of scoring functions for
  bayesian network model selection,'' in \emph{BMC bioinformatics}, vol.~13,
  no.~15.\hskip 1em plus 0.5em minus 0.4em\relax Springer, 2012, pp. 1--16.

\bibitem{kayaalp2012bayesian}
M.~Kayaalp and G.~F. Cooper, ``A bayesian network scoring metric that is based
  on globally uniform parameter priors,'' \emph{arXiv preprint
  arXiv:1301.0576}, 2012.

\bibitem{chen2008improving}
X.-W. Chen, G.~Anantha, and X.~Lin, ``Improving bayesian network structure
  learning with mutual information-based node ordering in the k2 algorithm,''
  \emph{IEEE Transactions on Knowledge and Data Engineering}, vol.~20, no.~5,
  pp. 628--640, 2008.

\bibitem{carvalho2009scoring}
A.~M. Carvalho, ``Scoring functions for learning bayesian networks,''
  \emph{Inesc-id Tec. Rep}, vol.~12, pp. 1--48, 2009.

\bibitem{jiang2011learning}
X.~Jiang, R.~E. Neapolitan, M.~M. Barmada, and S.~Visweswaran, ``Learning
  genetic epistasis using bayesian network scoring criteria,'' \emph{BMC
  bioinformatics}, vol.~12, no.~1, pp. 1--12, 2011.

\bibitem{ramsey2017million}
J.~Ramsey, M.~Glymour, R.~Sanchez-Romero, and C.~Glymour, ``A million variables
  and more: the fast greedy equivalence search algorithm for learning
  high-dimensional graphical causal models, with an application to functional
  magnetic resonance images,'' \emph{International journal of data science and
  analytics}, vol.~3, no.~2, pp. 121--129, 2017.

\bibitem{scheines2016measurement}
R.~Scheines and J.~Ramsey, ``Measurement error and causal discovery,'' in
  \emph{CEUR workshop proceedings}, vol. 1792.\hskip 1em plus 0.5em minus
  0.4em\relax NIH Public Access, 2016, p.~1.

\bibitem{zhang2012identifiability}
K.~Zhang and A.~Hyvarinen, ``On the identifiability of the post-nonlinear
  causal model,'' \emph{arXiv preprint arXiv:1205.2599}, 2012.

\bibitem{zhang2015estimation}
K.~Zhang, Z.~Wang, J.~Zhang, and B.~Sch{\"o}lkopf, ``On estimation of
  functional causal models: general results and application to the
  post-nonlinear causal model,'' \emph{ACM Transactions on Intelligent Systems
  and Technology (TIST)}, vol.~7, no.~2, pp. 1--22, 2015.

\bibitem{janzing2012information}
D.~Janzing, J.~Mooij, K.~Zhang, J.~Lemeire, J.~Zscheischler, P.~Daniu{\v{s}}is,
  B.~Steudel, and B.~Sch{\"o}lkopf, ``Information-geometric approach to
  inferring causal directions,'' \emph{Artificial Intelligence}, vol. 182, pp.
  1--31, 2012.

\bibitem{cai2020fom}
R.~Cai, J.~Ye, J.~Qiao, H.~Fu, and Z.~Hao, ``Fom: Fourth-order moment based
  causal direction identification on the heteroscedastic data,'' \emph{Neural
  Networks}, vol. 124, pp. 193--201, 2020.

\bibitem{jabbari2020instance}
F.~Jabbari and G.~F. Cooper, ``An instance-specific algorithm for learning the
  structure of causal bayesian networks containing latent variables,'' in
  \emph{Proceedings of the 2020 SIAM International Conference on Data
  Mining}.\hskip 1em plus 0.5em minus 0.4em\relax SIAM, 2020, pp. 433--441.

\bibitem{wang2018analysis}
Y.~Wang, ``Analysis of the max-min hill-climbing algorithm,'' in \emph{2018
  International Conference on Transportation \& Logistics, Information \&
  Communication, Smart City (TLICSC 2018)}.\hskip 1em plus 0.5em minus
  0.4em\relax Atlantis Press, 2018, pp. 509--511.

\bibitem{raghu2018comparison}
V.~K. Raghu, J.~D. Ramsey, A.~Morris, D.~V. Manatakis, P.~Sprites, P.~K.
  Chrysanthis, C.~Glymour, and P.~V. Benos, ``Comparison of strategies for
  scalable causal discovery of latent variable models from mixed data,''
  \emph{International journal of data science and analytics}, vol.~6, no.~1,
  pp. 33--45, 2018.

\bibitem{deng2013fine}
J.~Deng, J.~Krause, and L.~Fei-Fei, ``Fine-grained crowdsourcing for
  fine-grained recognition,'' in \emph{Proceedings of the IEEE conference on
  computer vision and pattern recognition}, 2013, pp. 580--587.

\bibitem{duan2012discovering}
K.~Duan, D.~Parikh, D.~Crandall, and K.~Grauman, ``Discovering localized
  attributes for fine-grained recognition,'' in \emph{2012 IEEE conference on
  computer vision and pattern recognition}.\hskip 1em plus 0.5em minus
  0.4em\relax IEEE, 2012, pp. 3474--3481.

\bibitem{krause2014learning}
J.~Krause, T.~Gebru, J.~Deng, L.-J. Li, and L.~Fei-Fei, ``Learning features and
  parts for fine-grained recognition,'' in \emph{2014 22nd International
  Conference on Pattern Recognition}.\hskip 1em plus 0.5em minus 0.4em\relax
  IEEE, 2014, pp. 26--33.

\bibitem{wah2011caltech}
C.~Wah, S.~Branson, P.~Welinder, P.~Perona, and S.~Belongie, ``The caltech-ucsd
  birds-200-2011 dataset,'' 2011.

\bibitem{mudambi2010research}
S.~M. Mudambi and D.~Schuff, ``Research note: What makes a helpful online
  review? a study of customer reviews on amazon. com,'' \emph{MIS quarterly},
  pp. 185--200, 2010.

\bibitem{luca2016reviews}
M.~Luca, ``Reviews, reputation, and revenue: The case of yelp. com,'' \emph{Com
  (March 15, 2016). Harvard Business School NOM Unit Working Paper}, no.
  12-016, 2016.

\bibitem{oghina2012predicting}
A.~Oghina, M.~Breuss, M.~Tsagkias, and M.~d. Rijke, ``Predicting imdb movie
  ratings using social media,'' in \emph{European conference on information
  retrieval}.\hskip 1em plus 0.5em minus 0.4em\relax Springer, 2012, pp.
  503--507.

\bibitem{tan2022causal}
F.~A. Tan, A.~H{\"u}rriyeto{\u{g}}lu, T.~Caselli, N.~Oostdijk, T.~Nomoto,
  H.~Hettiarachchi, I.~Ameer, O.~Uca, F.~F. Liza, and T.~Hu, ``The causal news
  corpus: Annotating causal relations in event sentences from news,''
  \emph{arXiv preprint arXiv:2204.11714}, 2022.

\bibitem{dror2017replicability}
R.~Dror, G.~Baumer, M.~Bogomolov, and R.~Reichart, ``Replicability analysis for
  natural language processing: Testing significance with multiple datasets,''
  \emph{Transactions of the Association for Computational Linguistics}, vol.~5,
  pp. 471--486, 2017.

\bibitem{zhang2015character}
X.~Zhang, J.~Zhao, and Y.~LeCun, ``Character-level convolutional networks for
  text classification,'' \emph{Advances in neural information processing
  systems}, vol.~28, 2015.

\bibitem{maliniak2013gender}
D.~Maliniak, R.~Powers, and B.~F. Walter, ``The gender citation gap in
  international relations,'' \emph{International Organization}, vol.~67, no.~4,
  pp. 889--922, 2013.

\bibitem{fu2013assessing}
K.-w. Fu, C.-h. Chan, and M.~Chau, ``Assessing censorship on microblogs in
  china: Discriminatory keyword analysis and the real-name registration
  policy,'' \emph{IEEE internet computing}, vol.~17, no.~3, pp. 42--50, 2013.

\bibitem{larson2019evaluation}
S.~Larson, A.~Mahendran, J.~J. Peper, C.~Clarke, A.~Lee, P.~Hill, J.~K.
  Kummerfeld, K.~Leach, M.~A. Laurenzano, L.~Tang \emph{et~al.}, ``An
  evaluation dataset for intent classification and out-of-scope prediction,''
  \emph{arXiv preprint arXiv:1909.02027}, 2019.

\bibitem{liu2019cm}
Y.~Liu, F.~Meng, J.~Zhang, J.~Zhou, Y.~Chen, and J.~Xu, ``Cm-net: A novel
  collaborative memory network for spoken language understanding,'' \emph{arXiv
  preprint arXiv:1909.06937}, 2019.

\bibitem{gould2022crisis}
M.~S. Gould, A.~Pisani, C.~Gallo, A.~Ertefaie, D.~Harrington, C.~Kelberman, and
  S.~Green, ``Crisis text-line interventions: Evaluation of texters'
  perceptions of effectiveness,'' \emph{Suicide and Life-Threatening Behavior},
  2022.

\bibitem{chen2019jddc}
M.~Chen, R.~Liu, L.~Shen, S.~Yuan, J.~Zhou, Y.~Wu, X.~He, and B.~Zhou, ``The
  jddc corpus: A large-scale multi-turn chinese dialogue dataset for e-commerce
  customer service,'' \emph{arXiv preprint arXiv:1911.09969}, 2019.

\bibitem{dasgupta2018automatic}
T.~Dasgupta, R.~Saha, L.~Dey, and A.~Naskar, ``Automatic extraction of causal
  relations from text using linguistically informed deep neural networks,'' in
  \emph{Proceedings of the 19th Annual SIGdial Meeting on Discourse and
  Dialogue}, 2018, pp. 306--316.

\bibitem{stukker2012subjectivity}
N.~Stukker and T.~Sanders, ``Subjectivity and prototype structure in causal
  connectives: A cross-linguistic perspective,'' \emph{Journal of pragmatics},
  vol.~44, no.~2, pp. 169--190, 2012.

\bibitem{ali2021causality}
W.~Ali, W.~Zuo, R.~Ali, X.~Zuo, and G.~Rahman, ``Causality mining in natural
  languages using machine and deep learning techniques: A survey,''
  \emph{Applied Sciences}, vol.~11, no.~21, p. 10064, 2021.

\bibitem{asghar2016automatic}
N.~Asghar, ``Automatic extraction of causal relations from natural language
  texts: a comprehensive survey,'' \emph{arXiv preprint arXiv:1605.07895},
  2016.

\bibitem{ben2011automatic}
A.~Ben~Abacha and P.~Zweigenbaum, ``Automatic extraction of semantic relations
  between medical entities: a rule based approach,'' \emph{Journal of
  biomedical semantics}, vol.~2, no.~5, pp. 1--11, 2011.

\bibitem{sharp2016creating}
R.~Sharp, M.~Surdeanu, P.~Jansen, P.~Clark, and M.~Hammond, ``Creating causal
  embeddings for question answering with minimal supervision,'' \emph{arXiv
  preprint arXiv:1609.08097}, 2016.

\bibitem{mirza2016catena}
P.~Mirza and S.~Tonelli, ``Catena: Causal and temporal relation extraction from
  natural language texts,'' in \emph{The 26th international conference on
  computational linguistics}.\hskip 1em plus 0.5em minus 0.4em\relax ACL, 2016,
  pp. 64--75.

\bibitem{abbott2016internet}
R.~Abbott, B.~Ecker, P.~Anand, and M.~Walker, ``Internet argument corpus 2.0:
  An sql schema for dialogic social media and the corpora to go with it,'' in
  \emph{Proceedings of the Tenth International Conference on Language Resources
  and Evaluation (LREC'16)}, 2016, pp. 4445--4452.

\bibitem{lopez2017discovering}
D.~Lopez-Paz, R.~Nishihara, S.~Chintala, B.~Scholkopf, and L.~Bottou,
  ``Discovering causal signals in images,'' in \emph{Proceedings of the IEEE
  conference on computer vision and pattern recognition}, 2017, pp. 6979--6987.

\bibitem{lv2022causality}
F.~Lv, J.~Liang, S.~Li, B.~Zang, C.~H. Liu, Z.~Wang, and D.~Liu, ``Causality
  inspired representation learning for domain generalization,'' in
  \emph{Proceedings of the IEEE/CVF Conference on Computer Vision and Pattern
  Recognition}, 2022, pp. 8046--8056.

\bibitem{zhang2021learning}
H.~Zhang, Y.~Huo, X.~Zhao, Y.~Song, and D.~Roth, ``Learning contextual
  causality between daily events from time-consecutive images,'' in
  \emph{Proceedings of the IEEE/CVF Conference on Computer Vision and Pattern
  Recognition}, 2021, pp. 1752--1755.

\bibitem{sisi2020finding}
T.~Sisi and Y.~Wan, ``Finding multiple variables causal dispositions in
  images.'' in \emph{ICCCS}, 2020, pp. 368--372.

\bibitem{barbu2019objectnet}
A.~Barbu, D.~Mayo, J.~Alverio, W.~Luo, C.~Wang, D.~Gutfreund, J.~Tenenbaum, and
  B.~Katz, ``Objectnet: A large-scale bias-controlled dataset for pushing the
  limits of object recognition models,'' \emph{Advances in neural information
  processing systems}, vol.~32, 2019.

\bibitem{buolamwini2018gender}
J.~Buolamwini and T.~Gebru, ``Gender shades: Intersectional accuracy
  disparities in commercial gender classification,'' in \emph{Conference on
  fairness, accountability and transparency}.\hskip 1em plus 0.5em minus
  0.4em\relax PMLR, 2018, pp. 77--91.

\bibitem{karkkainen2021fairface}
K.~Karkkainen and J.~Joo, ``Fairface: Face attribute dataset for balanced race,
  gender, and age for bias measurement and mitigation,'' in \emph{Proceedings
  of the IEEE/CVF Winter Conference on Applications of Computer Vision}, 2021,
  pp. 1548--1558.

\bibitem{hazirbas2021towards}
C.~Hazirbas, J.~Bitton, B.~Dolhansky, J.~Pan, A.~Gordo, and C.~C. Ferrer,
  ``Towards measuring fairness in ai: the casual conversations dataset,''
  \emph{IEEE Transactions on Biometrics, Behavior, and Identity Science}, 2021.

\bibitem{liu2018large}
Z.~Liu, P.~Luo, X.~Wang, and X.~Tang, ``Large-scale celebfaces attributes
  (celeba) dataset,'' \emph{Retrieved August}, vol.~15, no. 2018, p.~11, 2018.

\bibitem{kollias2018aff}
D.~Kollias and S.~Zafeiriou, ``Aff-wild2: Extending the aff-wild database for
  affect recognition,'' \emph{arXiv preprint arXiv:1811.07770}, 2018.

\bibitem{panda2018contemplating}
R.~Panda, J.~Zhang, H.~Li, J.-Y. Lee, X.~Lu, and A.~K. Roy-Chowdhury,
  ``Contemplating visual emotions: Understanding and overcoming dataset bias,''
  in \emph{Proceedings of the European Conference on Computer Vision (ECCV)},
  2018, pp. 579--595.

\bibitem{zheng2015person}
L.~Zheng, L.~Shen, L.~Tian, S.~Wang, J.~Bu, and Q.~Tian, ``Person
  re-identification meets image search,'' \emph{arXiv preprint
  arXiv:1502.02171}, 2015.

\bibitem{fu2017look}
J.~Fu, H.~Zheng, and T.~Mei, ``Look closer to see better: Recurrent attention
  convolutional neural network for fine-grained image recognition,'' in
  \emph{Proceedings of the IEEE conference on computer vision and pattern
  recognition}, 2017, pp. 4438--4446.

\bibitem{lou2019veri}
Y.~Lou, Y.~Bai, J.~Liu, S.~Wang, and L.~Duan, ``Veri-wild: A large dataset and
  a new method for vehicle re-identification in the wild,'' in
  \emph{Proceedings of the IEEE/CVF conference on computer vision and pattern
  recognition}, 2019, pp. 3235--3243.

\bibitem{hu2020probabilistic}
A.~Hu, F.~Cotter, N.~Mohan, C.~Gurau, and A.~Kendall, ``Probabilistic future
  prediction for video scene understanding,'' in \emph{European Conference on
  Computer Vision}.\hskip 1em plus 0.5em minus 0.4em\relax Springer, 2020, pp.
  767--785.

\bibitem{devlin2018bert}
J.~Devlin, M.-W. Chang, K.~Lee, and K.~Toutanova, ``Bert: Pre-training of deep
  bidirectional transformers for language understanding,'' \emph{arXiv preprint
  arXiv:1810.04805}, 2018.

\bibitem{xu2017scene}
D.~Xu, Y.~Zhu, C.~B. Choy, and L.~Fei-Fei, ``Scene graph generation by
  iterative message passing,'' in \emph{Proceedings of the IEEE conference on
  computer vision and pattern recognition}, 2017, pp. 5410--5419.

\bibitem{li2020molweni}
J.~Li, M.~Liu, M.-Y. Kan, Z.~Zheng, Z.~Wang, W.~Lei, T.~Liu, and B.~Qin,
  ``Molweni: A challenge multiparty dialogues-based machine reading
  comprehension dataset with discourse structure,'' \emph{arXiv preprint
  arXiv:2004.05080}, 2020.

\bibitem{busso2008iemocap}
C.~Busso, M.~Bulut, C.-C. Lee, A.~Kazemzadeh, E.~Mower, S.~Kim, J.~N. Chang,
  S.~Lee, and S.~S. Narayanan, ``Iemocap: Interactive emotional dyadic motion
  capture database,'' \emph{Language resources and evaluation}, vol.~42, no.~4,
  pp. 335--359, 2008.

\bibitem{poria2019emotion}
S.~Poria, N.~Majumder, R.~Mihalcea, and E.~Hovy, ``Emotion recognition in
  conversation: Research challenges, datasets, and recent advances,''
  \emph{IEEE Access}, vol.~7, pp. 100\,943--100\,953, 2019.

\bibitem{poria2018meld}
S.~Poria, D.~Hazarika, N.~Majumder, G.~Naik, E.~Cambria, and R.~Mihalcea,
  ``Meld: A multimodal multi-party dataset for emotion recognition in
  conversations,'' \emph{arXiv preprint arXiv:1810.02508}, 2018.

\bibitem{li2017dailydialog}
Y.~Li, H.~Su, X.~Shen, W.~Li, Z.~Cao, and S.~Niu, ``Dailydialog: A manually
  labelled multi-turn dialogue dataset,'' \emph{arXiv preprint
  arXiv:1710.03957}, 2017.

\bibitem{wang2020large}
Y.~Wang, P.~Ke, Y.~Zheng, K.~Huang, Y.~Jiang, X.~Zhu, and M.~Huang, ``A
  large-scale chinese short-text conversation dataset,'' in \emph{CCF
  International Conference on Natural Language Processing and Chinese
  Computing}.\hskip 1em plus 0.5em minus 0.4em\relax Springer, 2020, pp.
  91--103.

\bibitem{wu2016sequential}
Y.~Wu, W.~Wu, C.~Xing, M.~Zhou, and Z.~Li, ``Sequential matching network: A new
  architecture for multi-turn response selection in retrieval-based chatbots,''
  \emph{arXiv preprint arXiv:1612.01627}, 2016.

\bibitem{zhou2021eva}
H.~Zhou, P.~Ke, Z.~Zhang, Y.~Gu, Y.~Zheng, C.~Zheng, Y.~Wang, C.~H. Wu, H.~Sun,
  X.~Yang \emph{et~al.}, ``Eva: An open-domain chinese dialogue system with
  large-scale generative pre-training,'' \emph{arXiv preprint
  arXiv:2108.01547}, 2021.

\bibitem{welivita2021large}
A.~Welivita, Y.~Xie, and P.~Pu, ``A large-scale dataset for empathetic response
  generation,'' in \emph{Proceedings of the 2021 Conference on Empirical
  Methods in Natural Language Processing}, 2021, pp. 1251--1264.

\bibitem{srivastava2022dictionary}
S.~Srivastava and M.~Rohella, ``Dictionary vectorized hashing of emotional
  recognition of text in mutual conversation,'' in \emph{Artificial
  Intelligence and Speech Technology: Third International Conference, AIST
  2021, Delhi, India, November 12--13, 2021, Revised Selected Papers}.\hskip
  1em plus 0.5em minus 0.4em\relax Springer Nature, 2022, p. 215.

\bibitem{sinha2014investigating}
T.~Sinha and I.~Rajasingh, ``Investigating substructures in goal oriented
  online communities: Case study of ubuntu irc,'' in \emph{2014 IEEE
  International Advance Computing Conference (IACC)}.\hskip 1em plus 0.5em
  minus 0.4em\relax IEEE, 2014, pp. 916--922.

\bibitem{turcan2019dreaddit}
E.~Turcan and K.~McKeown, ``Dreaddit: A reddit dataset for stress analysis in
  social media,'' \emph{arXiv preprint arXiv:1911.00133}, 2019.

\bibitem{zhou2018emotional}
H.~Zhou, M.~Huang, T.~Zhang, X.~Zhu, and B.~Liu, ``Emotional chatting machine:
  Emotional conversation generation with internal and external memory,'' in
  \emph{Proceedings of the AAAI Conference on Artificial Intelligence},
  vol.~32, no.~1, 2018.

\bibitem{shen2021directed}
W.~Shen, S.~Wu, Y.~Yang, and X.~Quan, ``Directed acyclic graph network for
  conversational emotion recognition,'' \emph{arXiv preprint arXiv:2105.12907},
  2021.

\bibitem{allen1995natural}
J.~Allen, \emph{Natural language understanding}.\hskip 1em plus 0.5em minus
  0.4em\relax Benjamin-Cummings Publishing Co., Inc., 1995.

\bibitem{liu2019multi}
X.~Liu, P.~He, W.~Chen, and J.~Gao, ``Multi-task deep neural networks for
  natural language understanding,'' \emph{arXiv preprint arXiv:1901.11504},
  2019.

\bibitem{jhuang2013towards}
H.~Jhuang, J.~Gall, S.~Zuffi, C.~Schmid, and M.~J. Black, ``Towards
  understanding action recognition,'' in \emph{Proceedings of the IEEE
  international conference on computer vision}, 2013, pp. 3192--3199.

\bibitem{poppe2010survey}
R.~Poppe, ``A survey on vision-based human action recognition,'' \emph{Image
  and vision computing}, vol.~28, no.~6, pp. 976--990, 2010.

\bibitem{campbell1997speaker}
J.~P. Campbell, ``Speaker recognition: A tutorial,'' \emph{Proceedings of the
  IEEE}, vol.~85, no.~9, pp. 1437--1462, 1997.

\bibitem{peskin2003using}
B.~Peskin, J.~Navratil, J.~Abramson, D.~Jones, D.~Klusacek, D.~A. Reynolds, and
  B.~Xiang, ``Using prosodic and conversational features for high-performance
  speaker recognition: Report from jhu ws'02,'' in \emph{2003 IEEE
  International Conference on Acoustics, Speech, and Signal Processing, 2003.
  Proceedings.(ICASSP'03).}, vol.~4.\hskip 1em plus 0.5em minus 0.4em\relax
  IEEE, 2003, pp. IV--792.

\bibitem{sun2018emotional}
X.~Sun, X.~Chen, Z.~Pei, and F.~Ren, ``Emotional human machine conversation
  generation based on seqgan,'' in \emph{2018 First Asian Conference on
  Affective Computing and Intelligent Interaction (ACII Asia)}.\hskip 1em plus
  0.5em minus 0.4em\relax IEEE, 2018, pp. 1--6.

\bibitem{peng2019topic}
Y.~Peng, Y.~Fang, Z.~Xie, and G.~Zhou, ``Topic-enhanced emotional conversation
  generation with attention mechanism,'' \emph{Knowledge-Based Systems}, vol.
  163, pp. 429--437, 2019.

\bibitem{kann2022open}
K.~Kann, A.~Ebrahimi, J.~Koh, S.~Dudy, and A.~Roncone, ``Open-domain dialogue
  generation: What we can do, cannot do, and should do next,'' in
  \emph{Proceedings of the 4th Workshop on NLP for Conversational AI}, 2022,
  pp. 148--165.

\bibitem{majumder2020mime}
N.~Majumder, P.~Hong, S.~Peng, J.~Lu, D.~Ghosal, A.~Gelbukh, R.~Mihalcea, and
  S.~Poria, ``Mime: Mimicking emotions for empathetic response generation,''
  \emph{arXiv preprint arXiv:2010.01454}, 2020.

\bibitem{li2020dialbert}
T.~Li, J.-C. Gu, X.~Zhu, Q.~Liu, Z.-H. Ling, Z.~Su, and S.~Wei, ``Dialbert: A
  hierarchical pre-trained model for conversation disentanglement,''
  \emph{arXiv preprint arXiv:2004.03760}, 2020.

\bibitem{williams2014dialog}
J.~D. Williams, M.~Henderson, A.~Raux, B.~Thomson, A.~Black, and
  D.~Ramachandran, ``The dialog state tracking challenge series,'' \emph{AI
  Magazine}, vol.~35, no.~4, pp. 121--124, 2014.

\bibitem{mrkvsic2015multi}
N.~Mrk{\v{s}}i{\'c}, D.~O. S{\'e}aghdha, B.~Thomson, M.~Ga{\v{s}}i{\'c}, P.-H.
  Su, D.~Vandyke, T.-H. Wen, and S.~Young, ``Multi-domain dialog state tracking
  using recurrent neural networks,'' \emph{arXiv preprint arXiv:1506.07190},
  2015.

\bibitem{roman2019holistic}
M.~A. Rom{\'a}n, A.~Pertusa, and J.~Calvo-Zaragoza, ``A holistic approach to
  polyphonic music transcription with neural networks,'' \emph{arXiv preprint
  arXiv:1910.12086}, 2019.

\bibitem{medeot2018structurenet}
G.~Medeot, S.~Cherla, K.~Kosta, M.~McVicar, S.~Abdallah, M.~Selvi,
  E.~Newton-Rex, and K.~Webster, ``Structurenet: Inducing structure in
  generated melodies.'' in \emph{ISMIR}, 2018, pp. 725--731.

\bibitem{vercoe2001folk}
B.~L. Vercoe, ``Folk music classification using hidden markov models,'' in
  \emph{Proceedings of the International Conference on Artificial
  Intelligence}, vol.~6, 2001.

\bibitem{hawthorne2018enabling}
C.~Hawthorne, A.~Stasyuk, A.~Roberts, I.~Simon, C.-Z.~A. Huang, S.~Dieleman,
  E.~Elsen, J.~Engel, and D.~Eck, ``Enabling factorized piano music modeling
  and generation with the maestro dataset,'' \emph{arXiv preprint
  arXiv:1810.12247}, 2018.

\bibitem{kong2020giantmidi}
Q.~Kong, B.~Li, J.~Chen, and Y.~Wang, ``Giantmidi-piano: A large-scale midi
  dataset for classical piano music,'' \emph{arXiv preprint arXiv:2010.07061},
  2020.

\bibitem{bittner2014medleydb}
R.~M. Bittner, J.~Salamon, M.~Tierney, M.~Mauch, C.~Cannam, and J.~P. Bello,
  ``Medleydb: A multitrack dataset for annotation-intensive mir research.'' in
  \emph{ISMIR}, vol.~14, 2014, pp. 155--160.

\bibitem{politowski2020dataset}
C.~Politowski, F.~Petrillo, G.~C. Ullmann, J.~de~Andrade~Werly, and Y.-G.
  Gu{\'e}h{\'e}neuc, ``Dataset of video game development problems,'' in
  \emph{Proceedings of the 17th International Conference on Mining Software
  Repositories}, 2020, pp. 553--557.

\bibitem{abu2016youtube}
S.~Abu-El-Haija, N.~Kothari, J.~Lee, P.~Natsev, G.~Toderici, B.~Varadarajan,
  and S.~Vijayanarasimhan, ``Youtube-8m: A large-scale video classification
  benchmark,'' \emph{arXiv preprint arXiv:1609.08675}, 2016.

\bibitem{soomro2012ucf101}
K.~Soomro, A.~R. Zamir, and M.~Shah, ``Ucf101: A dataset of 101 human actions
  classes from videos in the wild,'' \emph{arXiv preprint arXiv:1212.0402},
  2012.

\bibitem{murray2012ava}
N.~Murray, L.~Marchesotti, and F.~Perronnin, ``Ava: A large-scale database for
  aesthetic visual analysis,'' in \emph{2012 IEEE conference on computer vision
  and pattern recognition}.\hskip 1em plus 0.5em minus 0.4em\relax IEEE, 2012,
  pp. 2408--2415.

\bibitem{carreira2017quo}
J.~Carreira and A.~Zisserman, ``Quo vadis, action recognition? a new model and
  the kinetics dataset,'' in \emph{proceedings of the IEEE Conference on
  Computer Vision and Pattern Recognition}, 2017, pp. 6299--6308.

\bibitem{carreira2018short}
J.~Carreira, E.~Noland, A.~Banki-Horvath, C.~Hillier, and A.~Zisserman, ``A
  short note about kinetics-600,'' \emph{arXiv preprint arXiv:1808.01340},
  2018.

\bibitem{wei-etal-2020-effective}
\BIBentryALTinterwordspacing
P.~Wei, J.~Zhao, and W.~Mao, ``Effective inter-clause modeling for end-to-end
  emotion-cause pair extraction,'' in \emph{Proceedings of the 58th Annual
  Meeting of the Association for Computational Linguistics}.\hskip 1em plus
  0.5em minus 0.4em\relax Online: Association for Computational Linguistics,
  Jul. 2020, pp. 3171--3181. [Online]. Available:
  \url{https://aclanthology.org/2020.acl-main.289}
\BIBentrySTDinterwordspacing

\bibitem{chen-etal-2020-end}
\BIBentryALTinterwordspacing
Y.~Chen, W.~Hou, S.~Li, C.~Wu, and X.~Zhang, ``End-to-end emotion-cause pair
  extraction with graph convolutional network,'' in \emph{Proceedings of the
  28th International Conference on Computational Linguistics}.\hskip 1em plus
  0.5em minus 0.4em\relax Barcelona, Spain (Online): International Committee on
  Computational Linguistics, Dec. 2020, pp. 198--207. [Online]. Available:
  \url{https://aclanthology.org/2020.coling-main.17}
\BIBentrySTDinterwordspacing

\bibitem{shen-etal-2021-directed}
\BIBentryALTinterwordspacing
W.~Shen, S.~Wu, Y.~Yang, and X.~Quan, ``Directed acyclic graph network for
  conversational emotion recognition,'' in \emph{Proceedings of the 59th Annual
  Meeting of the Association for Computational Linguistics and the 11th
  International Joint Conference on Natural Language Processing (Volume 1: Long
  Papers)}.\hskip 1em plus 0.5em minus 0.4em\relax Online: Association for
  Computational Linguistics, Aug. 2021, pp. 1551--1560. [Online]. Available:
  \url{https://aclanthology.org/2021.acl-long.123}
\BIBentrySTDinterwordspacing

\bibitem{luo2020mg}
J.~Luo, X.~Yang, S.~Ji, and J.~Li, ``Mg-vae: deep chinese folk songs generation
  with specific regional styles,'' in \emph{Proceedings of the 7th Conference
  on Sound and Music Technology (SCMT)}.\hskip 1em plus 0.5em minus 0.4em\relax
  Springer, 2020, pp. 93--106.

\bibitem{li2019regional}
J.~Li, J.~Luo, J.~Ding, X.~Zhao, and X.~Yang, ``Regional classification of
  chinese folk songs based on crf model,'' \emph{Multimedia tools and
  applications}, vol.~78, no.~9, pp. 11\,563--11\,584, 2019.

\bibitem{lian2021decn}
Z.~Lian, B.~Liu, and J.~Tao, ``Decn: Dialogical emotion correction network for
  conversational emotion recognition,'' \emph{Neurocomputing}, vol. 454, pp.
  483--495, 2021.

\bibitem{chen2022learning}
H.~Chen, X.~Yang, and C.~Li, ``Learning a general clause-to-clause
  relationships for enhancing emotion-cause pair extraction,'' 2022.

\bibitem{adel2017learning}
T.~Adel and C.~P. De~Campos, ``Learning bayesian networks with incomplete data
  by augmentation,'' in \emph{Thirty-first aaai conference on artificial
  intelligence}, 2017.

\bibitem{wei1990monte}
G.~C. Wei and M.~A. Tanner, ``A monte carlo implementation of the em algorithm
  and the poor man's data augmentation algorithms,'' \emph{Journal of the
  American statistical Association}, vol.~85, no. 411, pp. 699--704, 1990.

\bibitem{geman1984stochastic}
S.~Geman and D.~Geman, ``Stochastic relaxation, gibbs distributions, and the
  bayesian restoration of images,'' \emph{IEEE Transactions on pattern analysis
  and machine intelligence}, no.~6, pp. 721--741, 1984.

\bibitem{wang2020causal}
Y.~Wang, V.~Menkovski, H.~Wang, X.~Du, and M.~Pechenizkiy, ``Causal discovery
  from incomplete data: a deep learning approach,'' \emph{arXiv preprint
  arXiv:2001.05343}, 2020.

\bibitem{kaya2019deep}
M.~Kaya and H.~{\c{S}}. Bilge, ``Deep metric learning: A survey,''
  \emph{Symmetry}, vol.~11, no.~9, p. 1066, 2019.

\bibitem{danielsson1980euclidean}
P.-E. Danielsson, ``Euclidean distance mapping,'' \emph{Computer Graphics and
  image processing}, vol.~14, no.~3, pp. 227--248, 1980.

\bibitem{de2000mahalanobis}
R.~De~Maesschalck, D.~Jouan-Rimbaud, and D.~L. Massart, ``The mahalanobis
  distance,'' \emph{Chemometrics and intelligent laboratory systems}, vol.~50,
  no.~1, pp. 1--18, 2000.

\bibitem{kulis2013metric}
B.~Kulis \emph{et~al.}, ``Metric learning: A survey,'' \emph{Foundations and
  Trends{\textregistered} in Machine Learning}, vol.~5, no.~4, pp. 287--364,
  2013.

\bibitem{bucher2016hard}
M.~Bucher, S.~Herbin, and F.~Jurie, ``Hard negative mining for metric learning
  based zero-shot classification,'' in \emph{European Conference on Computer
  Vision}.\hskip 1em plus 0.5em minus 0.4em\relax Springer, 2016, pp. 524--531.

\bibitem{wang2021understanding}
F.~Wang and H.~Liu, ``Understanding the behaviour of contrastive loss,'' in
  \emph{Proceedings of the IEEE/CVF conference on computer vision and pattern
  recognition}, 2021, pp. 2495--2504.

\bibitem{ge2018deep}
W.~Ge, ``Deep metric learning with hierarchical triplet loss,'' in
  \emph{Proceedings of the European Conference on Computer Vision (ECCV)},
  2018, pp. 269--285.

\bibitem{wang2017deep}
J.~Wang, F.~Zhou, S.~Wen, X.~Liu, and Y.~Lin, ``Deep metric learning with
  angular loss,'' in \emph{Proceedings of the IEEE international conference on
  computer vision}, 2017, pp. 2593--2601.

\bibitem{sohn2016improved}
K.~Sohn, ``Improved deep metric learning with multi-class n-pair loss
  objective,'' \emph{Advances in neural information processing systems},
  vol.~29, 2016.

\bibitem{oh2016deep}
H.~Oh~Song, Y.~Xiang, S.~Jegelka, and S.~Savarese, ``Deep metric learning via
  lifted structured feature embedding,'' in \emph{Proceedings of the IEEE
  conference on computer vision and pattern recognition}, 2016, pp. 4004--4012.

\bibitem{he2019deep}
Z.~He, C.~Jung, Q.~Fu, and Z.~Zhang, ``Deep feature embedding learning for
  person re-identification based on lifted structured loss,'' \emph{Multimedia
  Tools and Applications}, vol.~78, no.~5, pp. 5863--5880, 2019.

\bibitem{liu2010study}
Z.~Liu, X.~Lv, K.~Liu, and S.~Shi, ``Study on svm compared with the other text
  classification methods,'' in \emph{2010 Second international workshop on
  education technology and computer science}, vol.~1.\hskip 1em plus 0.5em
  minus 0.4em\relax IEEE, 2010, pp. 219--222.

\bibitem{sun2020circle}
Y.~Sun, C.~Cheng, Y.~Zhang, C.~Zhang, L.~Zheng, Z.~Wang, and Y.~Wei, ``Circle
  loss: A unified perspective of pair similarity optimization,'' in
  \emph{Proceedings of the IEEE/CVF Conference on Computer Vision and Pattern
  Recognition}, 2020, pp. 6398--6407.

\end{thebibliography}

\noindent\textbf{Hang Chen} received the B.S. degree from the School of Computer Science and Technology, Xi'an JiaoTong University, Xi'an, China, in 2020. 

He is currently working toward the Ph.D. degree at the School of Computer Science and Technology, Xi'an JiaoTong University, Xi'an, China. His research interests include causal effect and analysis, causal representation of deep learning and natural language processing.

\vspace{11pt}
\vspace{11pt}

\noindent\textbf{Keqing Du} received the B.S. degree from the School of Statistics, Tianjin University of Finance and Economics, Tianjin, China, in 2022.

She is currently working toward the Ph.D. degree at the School of Computer Science and Technology, Xi'an JiaoTong University, Xi'an, China. Her research interests include causal effect and analysis, affective causal modeling and computer vision.

\vspace{11pt}
\vspace{11pt}

\noindent\textbf{Xinyu Yang} received the B.S., M.S., and Ph.D. degrees from Xi'an Jiaotong
University, Xi'an, China, in 1995, 1997, and 2001, respectively.

He is currently a Professor with the School of Computer Science and
Technology, Xi'an Jiaotong University. His research interests include affective
computing, image recognition, and audio and video data processing. He was
awarded the title of “New Century Excellent Talent” by the Chinese Ministry
of Education in 2009. He is a member of ACM.

\vspace{11pt}
\vspace{11pt}

\noindent\textbf{Chenguang Li} received the B.S. degree from the School of Mathematics and Statistics, Xi'an Jiaotong University, Xi'an, China, in 2021.

He is currently working toward the Ph.D. degree at the School of Computer Science and Technology, Xi'an Jiaotong University, Xi'an, China. His research interests include missing data causal completion, causal effect and analysis and time series processing.

\end{document}